\documentclass[runningheads]{llncs}

 \usepackage{eccv}

\usepackage{eccvabbrv}

\usepackage{url}
\usepackage{graphicx}
\usepackage{amsmath}
\usepackage{amsfonts,amssymb}
\usepackage{booktabs}
\usepackage{gensymb}
\usepackage{multirow}
\usepackage[english]{babel}
\usepackage{setspace}
\usepackage{booktabs}
\usepackage{array}
\usepackage{tabularx}
\usepackage{calc}
\usepackage{makecell}
\usepackage{pifont}
\usepackage{colortbl}
\usepackage{upgreek}
\usepackage{textcomp}
\usepackage{gensymb}
\usepackage[section]{placeins}
\usepackage[accsupp]{axessibility}  % Improves PDF readability for those with disabilities.

\usepackage{orcidlink}

\begin{document}
% \title{3D Dense Captioning beyond Nouns: A Middleware for Autonomous Driving}
    \title{\textit{$\text{TOD}^3$Cap}: \underline{To}war\underline{d}s 3\underline{D} \underline{D}ense Captioning in Outdoor Scenes}

\titlerunning{TOD3Cap: \underline{To}war\underline{d}s 3\underline{D} \underline{D}ense Captioning in Outdoor Scenes}

\author{Bu Jin$^{1,2}$,
Yupeng Zheng$^{1,2\dagger}$,
Pengfei Li$^{3}$,
Weize Li$^{3}$,
Yuhang Zheng$^{4}$,
Sujie Hu$^{3}$,
Xinyu Liu$^{5}$,
Jinwei Zhu$^{3}$,
Zhijie Yan$^{3}$,
Haiyang Sun$^{2}$,
Kun Zhan$^{2}$,
Peng Jia$^{2}$,
Xiaoxiao Long$^{6}$,
Yilun Chen$^{3}$,
and Hao Zhao$^{3}$ 
}

% \thanks{a}

\authorrunning{Jin et al.}

\institute{
$^{1}$CASIA $^{2}$Li Auto  $^{3}$AIR, Tsinghua University  $^{4}$Beihang University  $^{5}$HKUST   $^{6}$HKU \\
\email{Contact: jinbu18@mails.ucas.ac.cn, zhengyupeng2022@ia.ac.cn \\
Page: https://jxbbb.github.io/TOD3Cap} \\
$\dagger$ indicates the corresponding author.
}

% \\
% \url{http://www.springer.com/gp/computer-science/lncs}
 \maketitle
 \begin{figure}[!ht]
  \centering
  \includegraphics[width=0.95\textwidth]{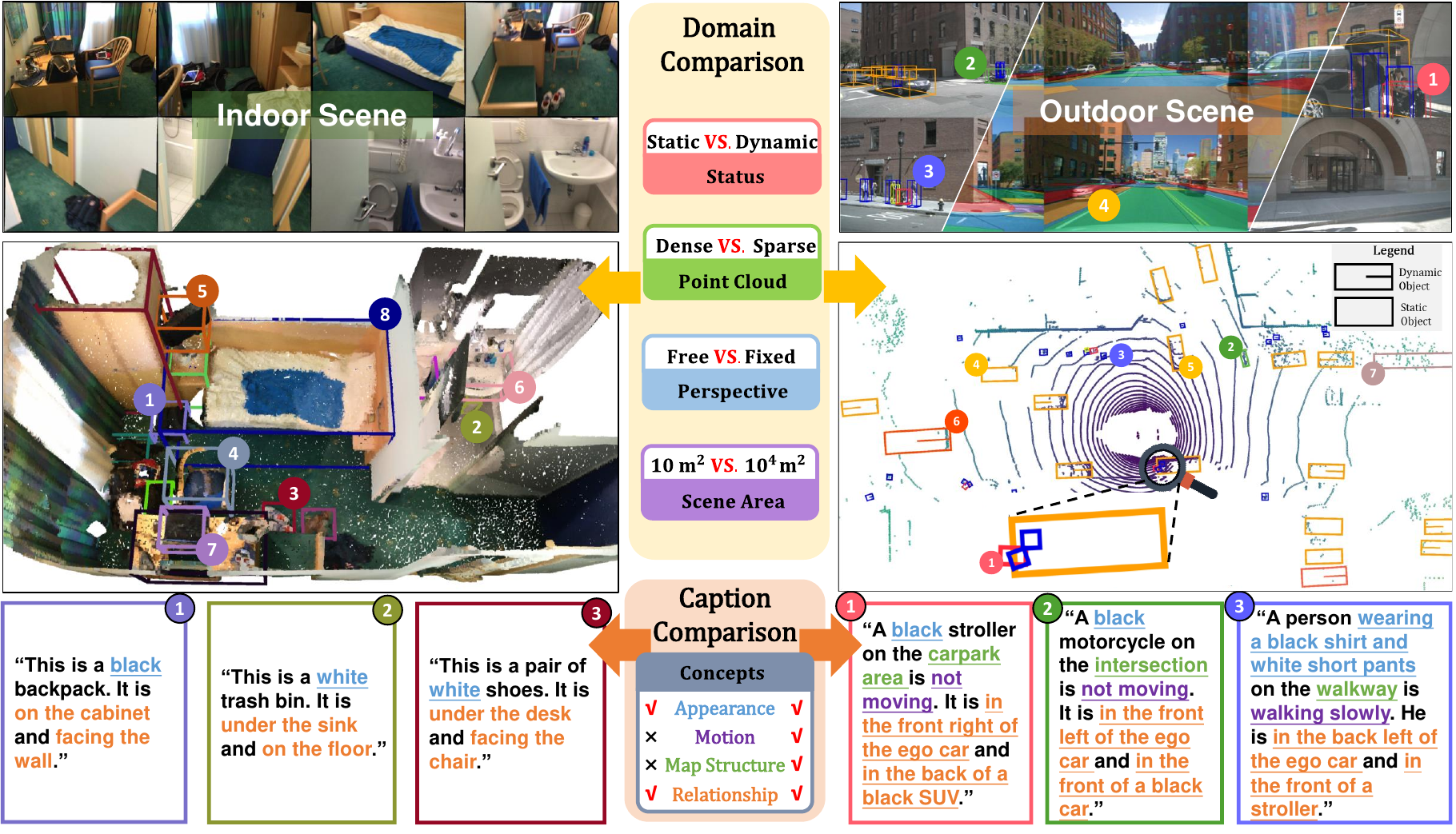}
  \caption{
  We introduce the task of 3D dense captioning in outdoor scenes (right). Given point clouds (right middle) and multi-view RGB inputs (right top), we predict box-caption pairs of all objects in a 3D outdoor scene. There are several fundamental domain gaps (middle column) between indoor and outdoor scenes, including \textcolor[RGB]{255,133,133}{{Status}}, \textcolor[RGB]{146,208,80}{{Point Cloud}}, \textcolor[RGB]{157,195,230}{{Perspective}}, and \textcolor[RGB]{180,130,218}{{Scene Area}}, bringing new challenges specific to outdoor scenes. Meanwhile, our outdoor 3D dense captioning (right bottom) contains more comprehensive concepts than indoor scenes (left bottom).
  }
  \label{fig:teaser}
\end{figure}
\vspace{-5pt}

% \footnote{The previous version of this paper can be found in \href{github首页}{https://github.com/}}

\begin{abstract}

3D dense captioning stands as a cornerstone in achieving a comprehensive understanding of 3D scenes through natural language. It has recently witnessed remarkable achievements, particularly in indoor settings.
However, the exploration of 3D dense captioning in outdoor scenes is hindered by two major challenges: 
1) the \textbf{domain gap} between indoor and outdoor scenes, such as dynamics and sparse visual inputs, makes it difficult to directly adapt existing indoor methods; 
2) the \textbf{lack of data} with comprehensive box-caption pair annotations specifically tailored for outdoor scenes.
To this end, we introduce the new task of outdoor 3D dense captioning. 
As input, we assume a LiDAR point cloud and a set of RGB images captured by the panoramic camera rig. The expected output is a set of object boxes with captions.
To tackle this task, we propose the \textbf{\textit{$\text{TOD}^3$Cap} network}, which leverages the BEV representation to generate object box proposals and integrates Relation Q-Former with LLaMA-Adapter to generate rich captions for these objects.
We also introduce the \textbf{\textit{$\text{TOD}^3$Cap} dataset}, the largest one to our knowledge for 3D dense captioning in outdoor scenes, which contains 2.3M descriptions of 64.3K outdoor objects from 850 scenes in nuScenes.
Notably, our \textit{$\text{TOD}^3$Cap} network can effectively localize and caption 3D objects in outdoor scenes, which outperforms baseline methods by a significant margin (+9.6 CiDEr@0.5IoU). Code, data and models are publicly available at \href{https://github.com/jxbbb/TOD3Cap}{https://github.com/jxbbb/TOD3Cap}.

  \keywords{3D dense captioning \and 3D scene understanding \and 3D vision language \and Dataset}
\end{abstract}

%------------------------------------------------------------------------------------------------%
\section{Introduction}

% - Background & application
Recently, the community has witnessed significant progress in 3D dense captioning. 
By explicitly formulating the understanding of 3D scenes with natural language, it exhibits diverse applications in cross-modal retrieval \cite{huang2018mhtn,chen2003visual}, robotic navigation \cite{savva2019habitat, huang2023visual, zhu2020vision, wang2019reinforced}, interactive AR/VR \cite{pidathala2022live} and autonomous driving \cite{sima2023drivelm, jin2023adapt, tian2024drivevlm}.
% By explicitly expressing its understanding of scenes, 3D dense captioning empowers machine to support a range of 3D vision-language learning tasks, like 3D visual grounding\cite{hong20223d, hsu2023ns3d, roh2022languagerefer} or 3D visual question answering\cite{azuma2022scanqa, ma2022sqa3d, etesam20223dvqa}. 
% Consequently, it opens up avenues for diverse applications such as cross-modal retrieval\cite{huang2018mhtn,chen2003visual}, robotic navigation\cite{savva2019habitat, huang2023visual, zhu2020vision, wang2019reinforced}, interactive AR/VR\cite{pidathala2022live}, and autonomous driving \cite{sima2023drivelm, jin2023adapt, tian2024drivevlm}.
% - talk about existing indoor approach with method and dataset
% So far, existing 3D dense captioning methods have been exclusively conducted on indoor scenes, benefiting from the strength of powerful 3D detectors, captioners, and datasets specifically developed for indoor environments.
In this challenging setting, an algorithm is required to localize all of the objects in a 3D scene and caption their diverse attributes.

% However, existing methods primarily focus on 3D indoor scenes, while outdoor 3D dense captioning has been relatively less explored.
% Upon revisiting the differences between indoor and outdoor scenes (shown in Fig.~\ref{fig:teaser}), the significant domain gap makes it impractical to directly apply indoor 3D dense captioning methods to outdoor settings. 

%3. introduce challenges
Some previous works \cite{chen2021scan2cap, chen2021d3net, yuan2022x, jiao2022more, cai20223djcg, chen2023vote2cap, wang2022spatiality} have explored the 3D dense captioning task and achieved promising results. 
However, these methods primarily focus on 3D dense captioning in indoor scenes, while outdoor 3D dense captioning has been rarely explored. Besides, taking a closer look at the significant differences between indoor and outdoor scenes (shown in Fig.~\ref{fig:teaser}), we argue it is sub-optimal to directly adapt these indoor methods to outdoor scenes, because:
\begin{itemize}
\item[$\bullet$] \textbf{Dynamic, not static.} Outdoor scenes are typically dynamic, necessitating the detection and tracking of objects with temporally changing status.
%\item[$\bullet$] \textbf{Less RGB images available.} While indoor scene scanning involves taking RGB images around a room with a high framerate, the average number of accessible RGB images for outdoor scenes results in a lack of visual cues, which can lead to the deficiency in appearance descriptions in the captioning. 
\item[$\bullet$] \textbf{Sparse LiDAR point clouds.} The utilization of sparse point clouds collected through LiDAR for outdoor scenes presents significant challenges in shape understanding. What's worse, the sparsity level is spatially varying. 
\item[$\bullet$] \textbf{Fixed camera perspective.} While indoor scene scanning allows free camera trajectories (e.g., around an object of interest), outdoor scenes typically feature a fixed 6-camera rig, presenting a higher degree of self-occlusion.
\item[$\bullet$] \textbf{Larger areas.} Outdoor scenes usually cover a significantly larger area.
\end{itemize}

\setlength{\tabcolsep}{2pt}
\begin{table}[t]
\begin{center}
\caption{Overview of existing 3D captioning datasets. The App., Mot., Env., Rel. denote Appearance, Motion, Environment and Relationship, respectively.
}
\label{tab:dataset_comparison}
\resizebox{1.0\textwidth}{!}
{
\setlength{\tabcolsep}{2.5mm}
\begin{tabular}{c|cc|cccc|ccc}
\Xhline{1.0pt}
\noalign{\smallskip}
Dataset & Domain & Dense Capt. & App. & Mot. & Env. & Rel. & $\sharp$Scenes &$\sharp$ Frames & $\sharp$Sentences\\
\noalign{\smallskip}
\Xhline{0.6pt}
\noalign{\smallskip}
Objaverse \cite{deitke2023objaverse} & Object &  \ding{55} & \ding{51} & \ding{55} & \ding{55} & \ding{55}  & - & N/A & 800K\\
\noalign{\smallskip}
\hline
\noalign{\smallskip}
SceneVerse \cite{jia2024sceneverse}  & Indoor & \ding{51} & \ding{51} & \ding{55} & \ding{51} & \ding{51} & 68K & N/A &2.5M\\
\noalign{\smallskip}
\hline
\noalign{\smallskip}
SceneFun3D \cite{delitzas2023scenefun3d}  & Indoor & \ding{55} & \ding{55} & \ding{51} & \ding{55} & \ding{51} & 710 & N/A &  14.8K \\
\noalign{\smallskip}
\hline
\noalign{\smallskip}
ScanRefer \cite{chen2020scanrefer} & Indoor & \ding{51} & \ding{51} & \ding{55} & \ding{55} & \ding{51} & 800 & N/A &  51.5K \\
\noalign{\smallskip}
\Xhline{0.6pt}
\noalign{\smallskip}
ReferIt3D \cite{achlioptas2020referit3d} & Indoor & \ding{51} & \ding{51} & \ding{55} & \ding{55} & \ding{51} & 800 & N/A &  41.5K \\
\noalign{\smallskip}
\hline
\noalign{\smallskip}
Multi3DRefer \cite{zhang2023multi3drefer} & Indoor & \ding{51} & \ding{51} & \ding{55} & \ding{55} & \ding{51} &  800 & N/A & 61.9K\\
\noalign{\smallskip}
\hline
\noalign{\smallskip}
nuCaption \cite{yang2023lidar} & \textbf{Outdoor} & \ding{55} & \ding{55} & \ding{51} & \ding{51} & \ding{51} & 265 & 34.1K & 420K\\
\noalign{\smallskip}
\hline
\noalign{\smallskip}
Rank2Tell \cite{sachdeva2024rank2tell} & \textbf{Outdoor} & \ding{55} & \ding{51} & \ding{51} & \ding{51} & \ding{51} & 116 & 5.8K &  -\\
\noalign{\smallskip}
\Xhline{0.6pt}
\hline
\noalign{\smallskip}
\rowcolor{gray!20}
\textbf{\textit{$\text{TOD}^3$Cap} (Ours)} & \textbf{Outdoor} & \ding{51} & \ding{51} & \ding{51} & \ding{51} & \ding{51} &  \textbf{850}  & \textbf{34.1K} &  \textbf{2.3M}\\
\noalign{\smallskip}
\Xhline{1.0pt}
\end{tabular}
}
\end{center}
\end{table}

These domain gaps pose significant challenges for successful 3D dense captioning in outdoor scenes. In this paper, we first formalize the new task of outdoor 3D dense captioning. It takes LiDAR point clouds and panoramic RGB images as inputs and the expected output is a set of object boxes with captions.
Then, we propose a transformer-based architecture, named \textit{$\text{TOD}^3$Cap} network, to address this task.
% The \textit{$\text{TOD}^3$Cap} network) 
Specifically, we first create a unified BEV map from 3D LiDAR point clouds and 2D multi-view images. Then we use a query-based detection head to generate object proposals. We also employ a Relation Q-Former to capture the relationship between object proposals and the scene context. The object proposal features are finally fed into a vision-language model to generate dense captions. Thanks to the usage of adapter\cite{zhang2023llama}, \textit{$\text{TOD}^3$Cap} network does not require re-training of the language model and thus we can leverage the commonsense in language foundation models pre-trained on a large corpus of data.

Apart from the fact that indoor-outdoor domain gaps render indoor architectures unsuitable for outdoor scenes, successfully addressing outdoor 3D dense captioning also suffers from the data hungriness \cite{yu2023comprehensive} issue, i.e., the lack of aligned box-caption pairs for outdoor scenes.
To facilitate future research in outdoor 3D dense captioning, we collect the \textit{$\text{TOD}^3$Cap} dataset, which provides box-wise natural language captions for LiDAR point cloud and panoramic RGB images from nuScenes\cite{caesar2020nuscenes}. In total, we acquire 2.3M captions of 63.4k outdoor instances. To the best of our knowledge, our \textit{$\text{TOD}^3$Cap} dataset is the first 3D dense captioning effort of million-scale sentences, the largest one to date for outdoor scenes. To summarize, our contributions are as follows:

\begin{itemize}
\item[$\bullet$] We introduce the outdoor 3D dense captioning task to densely detect and describe 3D objects, using LiDAR point clouds along with a set of panoramic RGB images as inputs. Its unique challenges are highlighted in Fig.~\ref{fig:teaser}.
\item[$\bullet$] We provide the \textit{$\text{TOD}^3$Cap} dataset containing 2.3M descriptions of 63.4k instances in outdoor scenes and adapt existing state-of-the-art approaches on our proposed \textit{$\text{TOD}^3$Cap} dataset for benchmarking.
% \item We propose \textit{$\text{TOD}^3$Cap} network a benchmark existing approaches on our \textit{$\text{TOD}^3$Cap} dataset.
\item[$\bullet$] We show that our method outperforms the baselines adapted from representative indoor methods by a significant margin (\textbf{+9.6 CiDEr@0.5IoU}).
\end{itemize}

%------------------------------------------------------------------------------------------------%

%------------------------------------------------------------------------------------------------%

\section{Related Work}

\subsubsection{3D Dense Captioning.}
Recently, the community has witnessed significant progress in 3D dense captioning \cite{chen2021scan2cap, chen2021d3net, yuan2022x, jiao2022more, cai20223djcg, chen2023vote2cap, wang2022spatiality}.
There are mainly two paradigms in previous research: ``detect-then-describe'' \cite{chen2021scan2cap, chen2021d3net, yuan2022x, jiao2022more, cai20223djcg, wang2022spatiality} and ``set-to-set''\cite{chen2023vote2cap}.
% Most existing methods  follow the ``detect-then-describe'' paradigm.
The ``detect-then-describe'' paradigm first utilizes a detector to generate proposals and then employs a generator to generate captions.
For example, Scan2Cap \cite{chen2021scan2cap} utilizes a VoteNet\cite{qi2019deep} to localize the objects in the scene, a graph-based relation module to model object relations and a decoder to generate sentences. \cite{chen2021d3net, chen2023unit3d} delve deeper to demonstrate the mutually reinforcing effect of dense captioning and visual grounding tasks. 
% \cite{wang2022spatiality} and \cite{jiao2022more} focus on improving 3D spatial relationships understanding.
Another approach to address the problem is the ``set-to-set'' paradigm, like Vote2Cap-DETR\cite{chen2023end} and its subsequent work\cite{chen2023vote2cap}. These methods treat the 3D dense captioning as a set-to-set problem and utilize the one-stage architecture to address it.  
% Our \textit{$\text{TOD}^3$Cap} follows the ``detect-then-describe'' paradigm.
% For concurrent work \cite{yang2023lidar} demonstrated the captioning ability using LiDAR feature in outdoor scenes, while our work further proposes new intermediate BEV features in the setting with sparse 3D information in outdoor scenes to achieve accurate and efficient 3D dense captioning.
Additionally, several works \cite{zhu20233d, hong20243d, chen2023ll3da, yang2023lidar} focus on large-scale pretraining by multitask settings to solve the 3D dense captioning task.
However, these methods are mainly focused on indoor scenarios and are difficult to adapt directly to outdoor scenes. In contrast, our proposed \textit{$\text{TOD}^3$Cap} network is aimed at outdoor 3D dense captioning.

\subsubsection{3D Captioning Datasets.}
Obtaining 3D language descriptions that are both object-centric and context-aware is a difficult job. Most commonly used datasets for 3D dense captioning are ScanRefer \cite{chen2020scanrefer} and ReferIt3D (Nr3D) \cite{achlioptas2020referit3d}, based on the richly-annotated 3D indoor dataset - Scannet \cite{dai2017scannet}. 
Notably, although recent developments like Objaverse\cite{deitke2023objaverse, deitke2024objaverse} have attempted large-scale object captioning for 3D-language alignment, they lack scene context information. Recently proposed indoor scene datasets like SceneVerse\cite{jia2024sceneverse}, SceneFun3D\cite{delitzas2023scenefun3d}, and Multi3DRefer\cite{zhang2023multi3drefer} focus on large-scale scene-graph captioning, object part-level captioning, and multi-object relationship captioning, respectively. 
However, existing datasets are mostly based on indoor scenes, which fail to cover unique scientific challenges of outdoor scenes as shown in Fig.~\ref{fig:teaser}. nuCaption\cite{yang2023lidar} and Rank2Tell\cite{sachdeva2024rank2tell} are designed for outdoor scenes, but they focus only on event-centric scene captioning instead of dense captioning. By contrast, our proposed \textit{$\text{TOD}^3$Cap} dataset provides dense object-centric language descriptions in outdoor scenes. We show the statistical comparison of our dataset with existing 3D captioning datasets in Tab.~\ref{tab:dataset_comparison}, highlighting its unique value.

\subsubsection{BEV-based 3D Perception.}
In recent years, there has been a rapid development and an increasing interest in BEV-based 3D perception techniques \cite{li2022bevformer, huang2021bevdet, li2023bevdepth, saha2022translating}, because BEV representation has proven to be highly beneficial for outdoor perception tasks such as 3D object detection and tracking. The Lift-Splat-Shoot \cite{philion2020lift} and its subsequent research\cite{hu2021fiery, li2023bevdepth} project image features into BEV pillar using predicted depth probabilities. BEVFormer\cite{li2022bevformer} utilizes a spatial cross attention to aggregate 2D image features into the BEV space and employs a temporal self attention to fuse temporal feature to model object motion. BEVFusion\cite{liu2023bevfusion} combines point cloud features from LiDAR and image features to enhance the geometric information in the BEV space. Inspired by them, our method fuses features from LiDAR and multi-view images and utilizes temporal fusion for obtaining richer contextual information and modeling object motion, which helps to address the challenges of outdoor dense captioning.

%------------------------------------------------------------------------------------------------%
% \section{Task}

%------------------------------------------------------------------------------------------------%
\section{\textit{$\text{TOD}^3$Cap} Dataset}
To facilitate research on outdoor 3D dense captioning task, we introduce \textit{$\text{TOD}^3$Cap}, a million-scale multi-modal dataset that extends the nuScenes \cite{caesar2020nuscenes} with dense captioning annotations. We introduce the data collection pipeline in Sec.~\ref{sec:data_collection}, and show the overall statistics of our proposed dataset in Sec.~\ref{sec:data_statistics}.

\subsection{Data Collection}
\label{sec:data_collection}

In this section, we introduce the data collection pipeline of the proposed dataset. We leverage a popular and large outdoor dataset nuScenes \cite{caesar2020nuscenes} encompassing 850 scenes for 3.4k frames. Each frame comprises 6 images taken from 6 cameras and point clouds from one LiDAR. The original dataset provides 3D bounding box annotations of 23 classes. We extend it to 3D dense captioning by annotating the appearance, motion, environment and relationship for all of the objects.

\subsubsection{Collection Principle.}
When describing an object in outdoor scenes, humans consider a series of questions \cite{cheng2023can}: ``What is it and what does it look like?'', ``What is it doing?'', ``Where is it?'', ``What is around it?'', which we refer to as their appearance, motion, environment and relationship, respectively. 
% We collect our data by asking the annotators to answer these questions.

\textbf{Appearance:}
The ability to describe what an object looks like is a hallmark of human intelligence. To answer the question, human annotators should recognize both the \textit{category} of the object and its \textit{visual attribute} (color, material, etc). 
For example, there is a person wearing blue shirts and black jeans.

\textbf{Motion:}
% What is it doing? Describing object actions represents human reasoning and predictions about the physical world. However, capturing the specific activities of an object from sparse LiDAR point clouds is extremely difficult. Thus, we simplify the modeling of object actions and mainly focus on the \textit{movement} of the object, like its speed or orientation. 
Different from the static indoor scenes, outdoor scenes are generally dynamic. In our annotation, we focus on the \textit{movement} of the object.
For example, a cat is moving away quickly or a dog is approaching slowly.

\textbf{Environment:}
% Although the 3D bounding box provides the exact location information of the object, humans do not use specific coordinate values to find an object. 
For outdoor scenes, an object's relative position in its environment is critical. So we ask the annotators to position the object roughly with its \textit{environment}. For example, there is a car in the parking lot.
% The map structure of an object represents its vertical proximity and refers to the spatial relationship between the object and its map locations.
% For example, a person is on the walkway or a car is in the stop lane.

\textbf{Relationship:}
Humans tend to find a \textit{reference} to describe an object, like ``the motorcycle next to the white truck'' or ``the stroller in the back left of the ego car''. 
% Thus it is also vital to model the spatial relations for the object when captioning. We provide the object-to-object relation for each target object. 
Following \cite{achlioptas2020referit3d}, we use the following compositional template for relation:
\begin{equation}
    \text{<target-object> <spatial-relation> <anchor-object>},
\end{equation}
where the target object represents the object to be described and the anchor object represents the anchor to describe the target.

\begin{figure*}[t]
  \centering
  \includegraphics[width=1.0\linewidth]{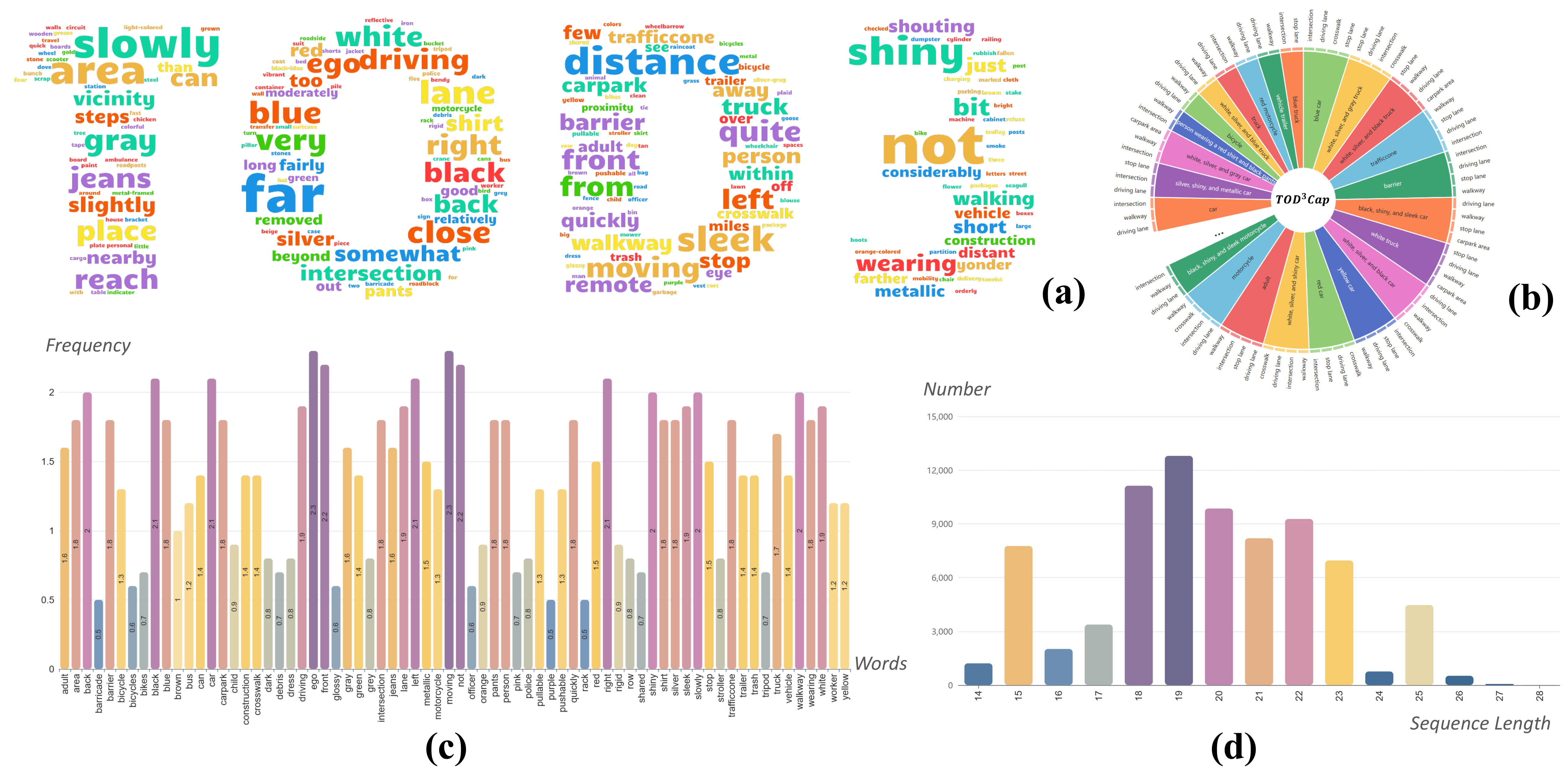}
  %\vspace{-2mm}
  \caption{The statistical properties of \textit{$\text{TOD}^3$Cap} dataset. (a) The word cloud visualization of \textit{$\text{TOD}^3$Cap}. (b) The visualization of object-environment relationship. (c) Statistics (in percentage) of the top 70 frequent words. (d) Sentence length Distribution. }
  
  %\vspace{-5mm}
  \label{fig:dataset_distribution}
\end{figure*}
\vspace{5pt}

\subsubsection{Annotation and Verification.}
With multiple annotation and validation steps, expert annotators make high-quality annotations for the object-level captions. 
% Specifically, the annotators annotate the four parts of the caption for all of the objects. 
% We first annotate the appearance, motion, and environment of the object, all of which represent the properties of the object itself. Then we annotate the object-to-object relation of the object. 

Notably, considering the significant success of large foundation models in language auto-labeling, we deploy a semi-automatic pipeline. Specifically, we first project the pre-labeled 3D bounding box to 2D. The 2D bounding box is then used to crop the camera image to an image patch that primarily consists of one object, which is then passed as input to a pre-trained captioning model (i.e., LLaMA-Adapter \cite{zhang2023llama}) to generate the captions for each object. Afterwards, we employ human annotators to perform strict correction and refinement of the generated sentences.  
% For the the object-to-object relation, we follow the annotation process in \cite{achlioptas2020referit3d} and synthesize the relation using the following compositional template:
% \begin{equation}
%     \text{<target-object> <spatial-relation> <anchor-object>}
% \end{equation}
% e.g., "It is closest to the man wearing blue shirts and black jeans". The human workers are required to find the anchor object that can uniquely characterize the target object. Note that the anchor object has a different class from the target. 
After labeling all of the four parts of the caption, we utilize GPT-4 \cite{ouyang2022training} to summarize them. Subsequently, the human workers are employed to check the correctness, fluency and readability of the captions. The annotation will not be reserved until three annotators reach an agreement.
We elaborate the details of the annotating process in the Appendix. 

\subsection{Data Statistics}
\label{sec:data_statistics}

In total, we employ ten expert human annotators to work for about 2000 hours. The total number of language descriptions is about 2.3M, with an average of 67.4 descriptions per frame and 2705.9 descriptions per scene.
We showcase the properties of our dataset in Fig.~\ref{fig:dataset_distribution}.
The descriptions cover over 500 types of outdoor objects with a total vocabulary of about 2k words. We find that the appearance of the object is generally more diverse than other attributes. The proportion of vocabulary for the appearance, motion, environment, and relationship is 69.7\%, 2.6\%, 7.1\%, and 20.6\%. Moreover, we find that humans use more words to describe the relations of objects. 
The average words of different parts are 3.7, 2.0, 2.9 and 11.2. Since our captions are very diverse and complex, successful dense captioning involves understanding the object-centric properties, object dynamics, object-object interactions, and object-environment relationships. More details about the dataset are provided in the appendix.

\section{\textit{$\text{TOD}^3$Cap} Network}
\begin{figure*}[t]
  \centering
  \includegraphics[width=0.95\linewidth]{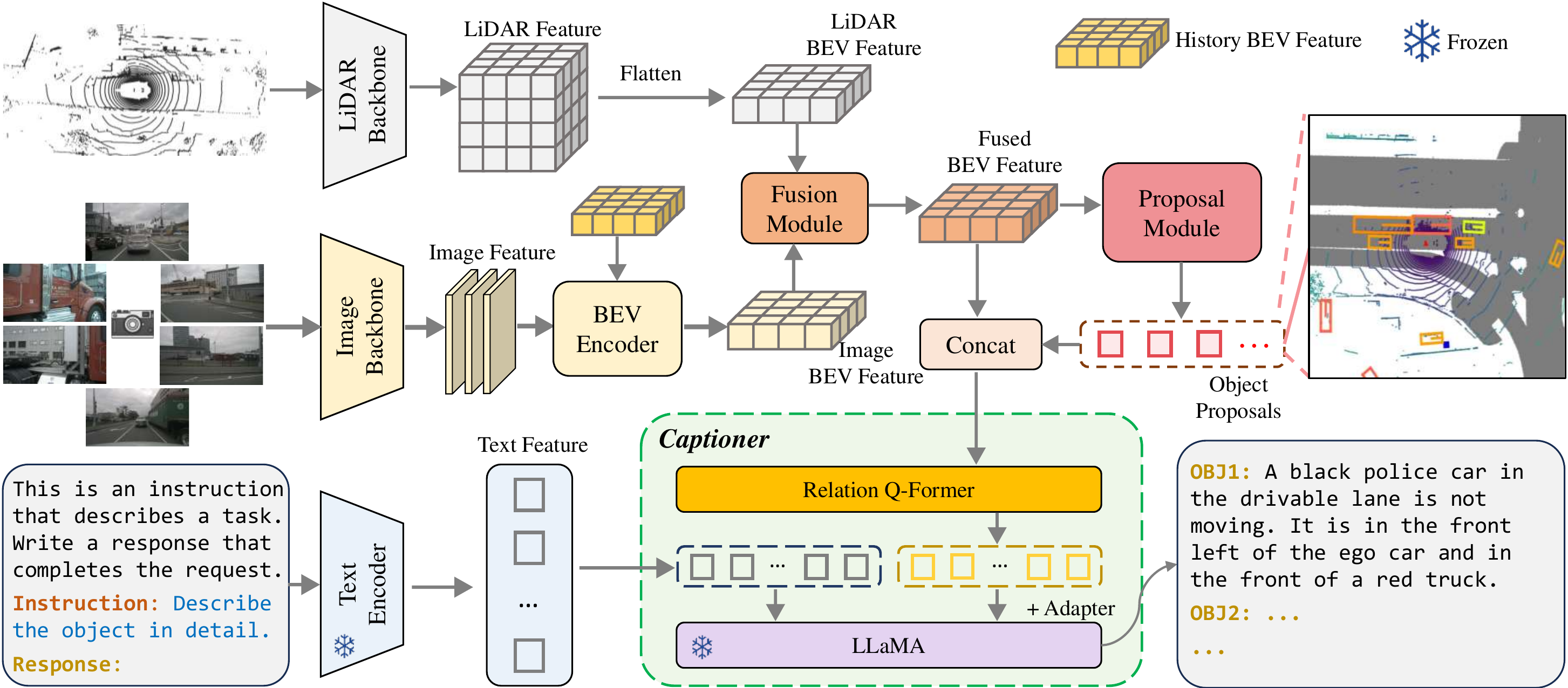}
  % \vspace{-2mm}
  \caption{\textbf{Architecture of our proposed \textit{$\text{TOD}^3$Cap} network.} Firstly, BEV features are extracted from 3D LiDAR point cloud and 2D multi-view images, followed by a query-based detection head that generates a set of 3D object proposals from the BEV features. Secondly, to capture the relationship information, we utilize a Relation Q-Former where the objects interact with other objects and the surrounding environment to get the context-aware features. Finally, with an Adapter, the features are processed to be prompts for the language model to generate dense captions. This formulation does not require a re-training process of the language model.}
  %% \vspace{-2mm}
  \label{fig:main}
  % \vspace{2mm}
\end{figure*}

To deal with the challenging outdoor 3D dense captioning problem, we propose a new end-to-end method named \textit{$\text{TOD}^3$Cap} network. An overview of \textit{$\text{TOD}^3$Cap} network architecture is shown in Fig.~\ref{fig:main}.

\textbf{Firstly}, BEV features are extracted from 3D LiDAR point cloud and 2D multi-view images, followed by a query-based detection head that generates a set of 3D object proposals from the BEV features (see Sec.~\ref{sec:bev_representation}).
\textbf{Secondly}, to capture the relationships between object proposals and scene context, we utilize a Relation Q-Former where the objects interact with other objects and the surrounding environment to get the context-aware features (see Sec. ~\ref{sec:relation_query_transformer}). 
\textbf{Finally}, with an Adapter \cite{zhang2023llama}, the object proposal features are processed to be prompts for the language model to generate dense captions. This formulation does not require a re-training process of the language model and thus we can leverage the commonsense of large foundation models pre-trained on a large corpus of data (see Sec.~\ref{sec:language_decoder}).

% % We introduce the task of dense driving captioning, the input of which is a driving scene with multi-view camera images and/or point cloud. Our goal is to design an architecture that can jointly localize the 3D bounding boxes for the underlying instances in the scene and generate corresponding detailed descriptions.
% As shown in Fig.~\ref{fig:main}, our goal is to jointly localize the 3D bounding boxes for the underlying instances and generate corresponding detailed descriptions. The input is a sequence of multi-view camera images $I = {I_1, I_2, ..., I_N}$ and the LiDAR point clouds $L$, where $N$ is the number of camera views. The expected output is a set of box-text pairs ($(B, C)={(b_1, c_1), ..., (b_n, c_n)}$), which represent the 3D positions and captions for $n$ objects in this scene.

% \vspace{-1mm}
\subsection{BEV-based Detector} \label{sec:bev_representation}
% \subsection{BEV Features and Object Proposals Generation} \label{sec:bev_representation}

Given multi-view camera images $I = \{I_i\}_{i=1}^N \in {\mathbb{R}}^{N \times H_c \times W_c \times 3}$ and LiDAR point clouds $L \in {\mathbb{R}}^{N_p \times 3}$, we first transform them into the unified BEV features $F_b \in {\mathbb{R}}^{H_b \times W_b \times C}$ and generate object proposals.

For multi-view images $I$, following \cite{li2022bevformer}, a spatial-temporal BEV encoder is used to lift image features to BEV space and effectively fuse the history BEV features to model dynamics.
Specifically, we first extract multi-view image features from $I$ with an image backbone. A set of learnable BEV queries $Q_c \in {\mathbb{R}}^{H_b \times W_b \times C}$ specific to camera are then updated by interacting with these features via spatial cross-attention layers \cite{li2022bevformer} to capture the spatial information, resulting in $F_c$:
\begin{align*} 
    F_{c} = \text{Spatial-Cross-Attention}(Q_c, \text{Backbone}(I)).
\end{align*} 
To model temporal dependency and capture dynamic features, if the preserved BEV features $F_c^p$ of the previous timestamp exist, the BEV queries $Q_c$ will first interact with $F_c^p$ through temporal self-attention layers, resulting in $Q_c^\prime$:
\begin{align*} 
    Q_c^\prime = \text{Temporal-Self-Attention}(Q_c, F_{c}^p).
\end{align*} 
For the initial timestamp, the BEV queries $Q_c$ are duplicated and fed into the temporal self-attention layers. The resulted $Q_c^\prime$ are then taken as the input of the spatial cross-attention layers as a substitute for $Q_c$.

For LiDAR input $L$, a LiDAR backbone is first employed to extract voxelized LiDAR features. Then, the features are flattened along the height dimension, leading to the BEV features $F_l \in {\mathbb{R}}^{H_b \times W_b \times C}$. 
Finally, the BEV features of the two different modalities are fused together with a convolutional fusion module to acquire the unified BEV features $F_{b}$.

Subsequently, we exploit a query-based object proposal generation module that takes the BEV features $F_b$ as input to generate the object box proposals $\hat{B} = \{\hat{B}_i\}_{i=1}^K \in {\mathbb{R}}^{K \times D}$, where $K$ is the preset number of object queries and $D$ corresponds to the dimension of proposal feature. The process of proposal generation aligns with that in traditional detection head like DETR \cite{carion2020end}. 

\vspace{-3mm}
\subsection{Relation Q-Former}
\label{sec:relation_query_transformer}

After obtaining the BEV features $F_b$ and object proposals $\hat{B}$, a relation query transformer (Relation Q-Former) is designed to extract context-aware features for each object.
% We tokenize the feature map $F_b^t \in {\mathbb{R}}^{H \times W \times C}$ along the channel dimension, resulting in $H \times W$ BEV tokens of size $C$.
Specifically, we first create object queries by encoding the object proposals $\hat{B}$ with a learnable MLP, resulting in object features with the same feature dimension as $F_b$. These features are then concatenated and fed into the Relation Q-Former, which comprises several self-attention layers for feature interaction. As shown later, this module improves performance significantly.
% This process could be formulated as follows:
\begin{align*} 
    Q_B &= \text{Relation Q-Former}(\text{MLP}(\hat{B}), F_b).
\end{align*} 
The resulting object queries $Q_B$ are taken as input to a captioning decoder for natural language generation, which will be elaborated in the next section.

% To avoid the memory cost and optimization difficulty in caption generation for $N$ proposals, we filter the queries by a confidence threshold and randomly sample the filtered results to $N_s$, which is much smaller than $N$. 

\subsection{Captioning Decoder}
\label{sec:language_decoder}
Inspired by the recent advancements of LLMs in contextual reasoning, we employ a frozen LLM as our language generator, which takes object queries $Q_B$ as input and output descriptions for each object. 
To ensure the dimension consistency between $Q_B$ and the hidden layers of the LLM, we first use an MLP to transform the dimension of $Q_B$, resulting in $Q_B^\prime$. We further employ an Adapter \cite{zhang2023llama} to align the object proposal representation with the feature space of the pre-trained language model, which bridges the modality gap. The adapted object features serve as prompts $\mathcal{V}$ for the LLM to generate corresponding captions. 
\begin{align*} 
    Q_B^\prime = \text{MLP}(Q_B),&\quad \mathcal{V} = \text{Adapter}(Q_B^\prime), \\
    \hat{\mathcal{C}} =& \text{LLM}(\mathcal{T}, \mathcal{V}),
\end{align*} 
where $\mathcal{T}$ is the system text prompt (as shown in the left-bottom corner of Fig.~\ref{fig:main}) and $\hat{\mathcal{C}} = \{\hat{w}_i\}_{i=1}^M$ is the resulting caption, which consists of $M$ words.
% where $\mathcal{T}$ is the system text prompt and $\hat{\mathcal{C}} = \{\hat{\mathcal{C}}_i\}_{i=1}^K$ is the set of resulting captions. Each caption of $i$-th predited object consists of $M$ words $\hat{\mathcal{C}}_i = \{\hat{w}_i^j\}_{j=1}^M$.

During training process, we take the standard cross-entropy loss as the captioning loss $\mathcal{L}_{\text{cap}}$ and train the model in the \emph{teacher-forcing}\footnote{It means using ground truth words as the conditioning during training, which differs from the auto-regressive testing setting that uses predicted words for conditioning.} manner:
\begin{equation}
\mathcal{L}_{\text{cap}}=\sum_{i=1}^M \mathcal{L}_{\text{cap}}(w_i)=-\sum_{i=1}^M \log \hat{p}\left(w_i \mid w_{[1: i-1]}, \mathcal{T}, \mathcal{V},\theta_\text{LLM} \right),
\end{equation}
where ${\mathcal{C}} = \{{w}_i\}_{i=1}^M$ is the ground truth caption, $\theta_\text{LLM}$ represent the weights of the LLM and $\hat{p}$ is the predicted probability.
Note that $\theta_\text{LLM}$ are frozen to reduce the computation cost and mitigate the catastrophic forgetting problem of LLM.

% Note that the LLM is frozen 

Moreover, considering the memory burden and optimization difficulty when generating hundreds of sentences during training, we do not feed all the object queries into the captioning decoder at once. Instead, we filter the queries by a 3D hungarian assigner \cite{wang2022detr3d} to get those matched with the ground truth and then randomly sample a subset during training. During inference, we apply non-maximum suppression (NMS) to suppress overlapping proposals.

\subsection{Loss Function}
We utilize $L_1$ loss as $\mathcal{L}_{\text{obj}}$ to supervise 3D bounding box regression for object proposal generation and use $\mathcal{L}_{\text{cap}}$ for captioning.
Then the overall loss for dense captioning is calculated as the weighted combination:
% \vspace{-0.5mm}
\begin{equation}
    \mathcal{L} = \alpha \mathcal{L}_{\text{obj}} + \beta \mathcal{L}_{\text{cap}}.
\end{equation}
where hyper-parameters $\alpha$ and $\beta$ are set to $\alpha=10$ and $\beta=1$ in our experiments.

\section{Experiments}
% To benchmark our proposed \textit{$\text{TOD}^3$Cap} network dataset and our \textit{$\text{TOD}^3$Cap} network baseline model, we evaluate existing indoor 3D dense captioning methods.

We conduct a comprehensive evaluation of adapted state-of-the-art baseline methods and ours on \textit{$\text{TOD}^3$Cap} dataset. In Sec.~\ref{sec:experimental_setup}, we describe the evaluation metrics, the implementation details of our model and adapted baselines. 
In Sec.~\ref{sec:baseline}, we compare adapted indoor baselines with our proposed method on the introduced dataset. Finally in Sec.~\ref{sec:ablation_study}, we conduct a comprehensive ablation study to validate the effectiveness of \textit{$\text{TOD}^3$Cap} network design.
% The experiments are conducted independently on two different settings: (1) 3D dense captioning that takes multi-view camera images and/or point cloud as input and generates 3D bounding boxes and captions; (2) 2D dense captioning that takes single image as input and generates 2D bounding boxes and captions. Intuitively, 3D captioning has more advantages in spatial awareness, while 2D captioning outperforms in visual recognition.

\subsection{Experimental Setup}
\label{sec:experimental_setup}
\subsubsection{Dataset and Metrics.}
We inherit the official nuScenes split setting for \textit{$\text{TOD}^3$Cap}, where the train/val scenes are 700 and 150, respectively. The reported results are calculated on the val split for all following experiments.
The $m@kIoU$ metric \cite{chen2021scan2cap} is leveraged for the evaluation of the 3D outdoor dense captioning task.
% We denote each predicted box-caption pair as $(\hat{B}_i, \hat{\mathcal{C}}_i)$ where $\hat{B}_i$ represents the bounding box of the $i$-th predicted object and $\hat{\mathcal{C}}_i$ represents the corresponding caption.
% The ground truth is denoted as $(B_i, \mathcal{C}_i)$ where $B_i$ is the bounding box label and $\mathcal{C}_i$ is the corpus of the ground truth captions. 
Specifically, we denote each ground truth box-caption pair as $(B_i, \mathcal{C}_i)$, where $B_i$ and $\mathcal{C}_i$ are the bounding box label and the ground truth caption for the $i$-th object. The predicted box-caption pair matched with the ground truth is denoted as $(\hat{B}_i, \hat{\mathcal{C}}_i)$.
For all $(B_i, \mathcal{C}_i)$ and $(\hat{B}_i, \hat{\mathcal{C}}_i)$, the m@kIoU is defined as:
% \vspace{-1mm}
\begin{equation}
m@kIoU=\frac{1}{N_{\text{gt}}} \sum_{i=1}^{N_{\text{gt}}} m\left(\hat{\mathcal{C}}_i, {\mathcal{C}}_i\right) \cdot \mathbb{I}\left\{\operatorname{IoU}\left(\hat{B}_i, B_i\right) \geq k\right\},
\end{equation}
where $N_{\text{gt}}$ is the number of the ground truth objects and  $m$ represents the standard image captioning metrics, including BLEU \cite{papineni2002bleu}, METEOR \cite{banerjee2005meteor}, Rouge \cite{lin2004rouge} and CIDEr \cite{vedantam2015cider}, abbreviated as B, M, R, C, respectively. 

\begin{table*}[!ht]
    \caption{Quantitative results on \textit{$\text{TOD}^3$Cap} dataset. The ``*'' represents that we replace the scene encoder with BEV encoder for adaptation. All of the methods are trained to full convergence on the \textit{$\text{TOD}^3$Cap} dataset for fair comparison. Our \textit{$\text{TOD}^3$Cap} network outperforms other methods with a clear margin, using various inputs. }
    \label{tab:comp_baseline}
    \centering
    \resizebox {\linewidth}{!}{
        \begin{tabular}{c|c|cccc|cccc}
             \toprule 
             Method & Input & C@0.25 & B-4@0.25 & M@0.25 & R@0.25 & C@0.5 & B-4@0.5 & M@0.5 & R@0.5 \\

            \midrule

            \textit{$\text{TOD}^3$Cap} (Ours)    & 2D      & 96.2 & 45.0 & 34.2 & 67.4 & 94.1 & 47.6 & 33.3 & 65.4 \\

            \midrule

            % Scan2Cap \cite{chen2021scan2cap}      & 3D & - & - & - & - & -  & - & - & - \\
            Scan2Cap* \cite{chen2021scan2cap}      & 3D & 50.6 &34.3 & 25.2 & 57.9 & 43.3  & 31.3 & 22.8 & 50.8 \\
            Vote2Cap-DETR* \cite{chen2023vote2cap}          & 3D & 72.8 & 41.6 & 29.5 & 60.6 &62.6  & 35.9 & 27.4 & 55.8 \\
             % \cite{chen2023vote2cap}  & {2D+3D} & {110.1} & {48.0} & {44.4} & {67.8} & {98.4} & {46.1} & {41.3} & {65.1} \\

            \textit{$\text{TOD}^3$Cap} (Ours)  & 3D & 85.3 & 43.0 & 29.9 & 60.5 & 74.4 & 39.4 & 27.2 & 55.4       \\

            \midrule
            % Scan2Cap \cite{chen2021scan2cap}      & 2D+3D & - & - & - & - & -  & - & - & - \\
            Scan2Cap* \cite{chen2021scan2cap}     & 2D+3D & 60.6 & 41.5 & 28.4 & 58.6 & 62.5 & 39.2 & 26.4 & 56.5 \\
            X-Trans2Cap* \cite{yuan2022x}   & 2D+3D & 99.8 & 45.9 & 35.5 & 66.8 & 92.2 & 43.3 & 34.7 & 65.7 \\
            Vote2Cap-DETR* \cite{chen2023vote2cap}  & {2D+3D} & {110.1} & {48.0} & {44.4} & {67.8} & {98.4} & {46.1} & {41.3} & {65.1} \\

            \textit{$\text{TOD}^3$Cap} network (Ours)  & 2D+3D     & \textbf{120.3} & \textbf{51.5} & \textbf{45.1} & \textbf{70.1} & \textbf{108.0} & \textbf{50.2} & \textbf{48.9} & \textbf{69.2} \\
            \bottomrule
        \end{tabular}
    }
\end{table*}

\subsubsection{Baselines.}
% describe how to adapt the existing indoor method to the outdoor dataset.
From existing methods for 3D dense captioning, we take milestone methods and state-of-the-art methods \cite{chen2021scan2cap, yuan2022x, chen2023vote2cap} for benchmarking:
% Scan2Cap
% These arts typically consist of three principal components: scene encoder, relation module, and feature decoder. 
(1) Scan2Cap \cite{chen2021scan2cap} utilizes the VoteNet\cite{qi2019deep} detector to localize objects in a scene and uses a graph-based relation module to explore object relations. (2) X-Trans2Cap \cite{yuan2022x} utilizes a teacher-student framework to transfer the rich appearance information from 2D images to 3D scenes. (3) Vote2Cap-DETR \cite{chen2023vote2cap} adopts a one-stage architecture that applies two parallel prediction heads to decode the scene features into bounding boxes and the corresponding captions.

\subsubsection{Adaptation.} These methods involve domain-specific design choices for 3D indoor scenes. However, directly applying them to outdoor scenes leads to sub-optimal performance. A major challenge is that their detectors cannot effectively locate outdoor objects because of the varying sparsity of LiDAR point clouds and the limited number of camera viewpoints.
For a fair comparison, we adapt these methods to the outdoor setting by (1) replacing their detector with the same one as ours and (2) loading our pre-trained detector weights. In this way, these methods obtain the same localization capabilities as ours. All these methods are then trained on the \textit{$\text{TOD}^3$Cap} dataset until convergence. In Tab.\ref{tab:comp_baseline}, adapted baseline methods are marked with *. The comparison between baseline methods before and after adaptation is provided in the appendix (Tab.~\ref{tab:comp_adaptation}), showing a substantial upgrade.

% describe our \textit{$\text{TOD}^3$Cap} implementation details.

\subsubsection{Protocol.} For the proposed \textit{$\text{TOD}^3$Cap} network, we train the network in three stages to facilitate the optimization process.
Firstly, the BEV-based detector is pre-trained on object detection task. We train the detector on the train split of nuScenes with 24 epochs and a learning rate of 2e-4. 
Then the weights of the BEV-based detector are frozen and the object box proposals are utilized to generate captions. We train this stage with 10 epochs and a learning rate of 2e-4.
Finally, the entire model is finetuned with a smaller learning rate of 2e-5 for 10 epochs. 
We employ AdamW \cite{loshchilov2017decoupled} with a weight decay of 1e-2 as the optimizer.
The pre-trained LLaMA-7B \cite{zhang2023llama} is taken as the LLM in our captioning decoder.

\subsection{Comparing with State-of-the-art Methods}
\label{sec:baseline}
\subsubsection{Quantitative Results.}
We show results separately for different input modalities, including (1) multi-view RGB images (denoted as 2D), (2) LiDAR point clouds (denoted as 3D), and (3) both images and point clouds (denoted as 2D+3D). The quantitative results are shown in Table.~\ref{tab:comp_baseline}, demonstrating:
% Depending on the input modality, we exploit different BEV feature extractors: BEVFormer \cite{li2022bevformer}, Centerpoint \cite{yin2021center} and BEVFusion \cite{liu2023bevfusion}.

\noindent \textbf{(1) \textit{$\text{TOD}^3$Cap} network outperforms prior arts.} 
Specifically, when taking 2D images and 3D point clouds (2D+3D) as input, the proposed \textit{$\text{TOD}^3$Cap} network outperforms Vote2Cap-DETR by 10.2 (9.26\%) on C@0.25 and 9.6 (9.76\%) on C@0.5. When taking only point clouds as input, our \textit{$\text{TOD}^3$Cap} network achieves 12.5 (17.17\%) and 11.8 (18.85\%) improvement over Vote2Cap-DETR.
These results indicates the effectiveness of the proposed \textit{$\text{TOD}^3$Cap} network.

\noindent \textbf{(2) The multi-modal input improves captioning performance.}
The performance of \textit{$\text{TOD}^3$Cap} network with multi-modal input outperforms that with the camera-only or LiDAR-only input, indicating that the information from the camera and LiDAR are complementary to each other. 
For LiDAR-only results, the sparsity of LiDAR point clouds makes it challenging to capture the visual attributes and textures of objects.
For camera-only results, it is difficult to capture distance information of objects solely based on images, which results in the poor captioning related to motion and environment. We provide qualitative results in the appendix to support the complementary nature of multi-modal inputs.
% \jb{TODO: add a figure to prove this.}

% \noindent \textbf{(3) BEV representations enable indoor methods to be applied in outdoor scenes.}
% We can see that the direct application of the indoor methods on DESIGN results a failure during training. However, when we replace the intermediate representation by BEV, the performance will be greatly improved.

\subsubsection{Qualitative Analysis.}
\label{sec:qualitative_analysis}
\begin{figure*}[t]
  \centering
  \includegraphics[width=1.0\linewidth]{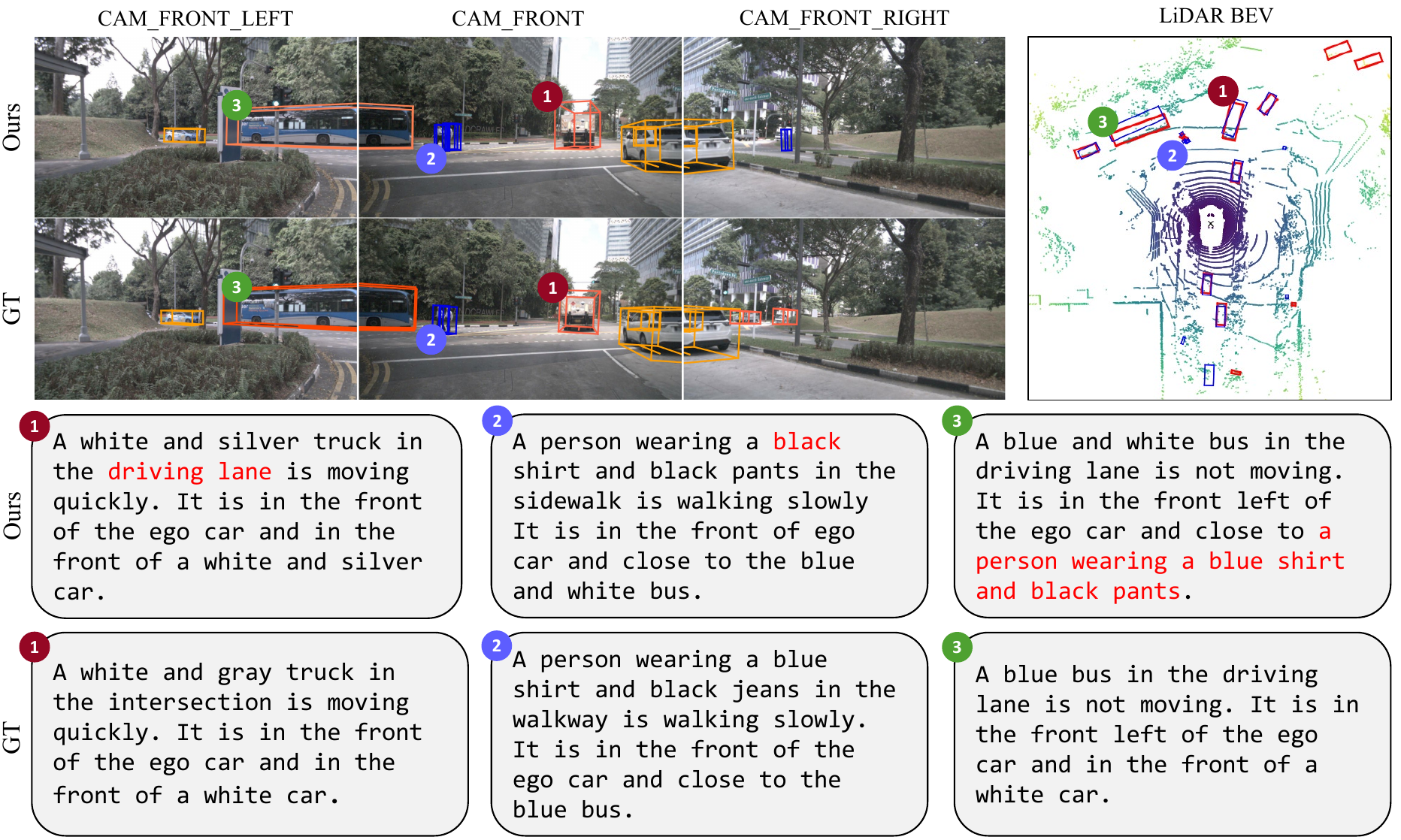}
  % \vspace{2mm}
  \caption{Qualitative results for our proposed \textit{$\text{TOD}^3$Cap} network. In the top left, we show our predicted bounding boxes and corresponding captions in the first row and ground truth in the second row. In the top right, we show our predicted bounding boxes in \textcolor[RGB]{0,0,255}{blue} and the ground truth bounding boxes in \textcolor[RGB]{255,0,0}{red}. In the bottom, we mark the wrong descriptions in \textcolor[RGB]{255,0,0}{red}.
  The \textit{$\text{TOD}^3$Cap} network produces impressive results except for a few mistakes.}
  % \vspace{2mm}
  \label{fig:qualitative}
\end{figure*}

We show some qualitative results in Fig.~\ref{fig:qualitative}, including the detection results and corresponding descriptions. We can see \textit{$\text{TOD}^3$Cap} network accurately localizes most objects and provides sound descriptions, except for a few mistakes in small and remote objects, calling for future algorithmic development in this novel and important outdoor 3D dense captioning problem. 
% For the remote objects, we can see that the model can not 
% We observe that \textit{$\text{TOD}^3$Cap} network is able to recognize the rare properties (or long-tailed properties), which demonstrates its superior open scene understanding ability. For example, a detected person is indeed a construction worker (the second item of the predictions in Fig.~\ref{fig:qualitative}). More results can be found in the appendix.

\subsection{Ablation Study}
\label{sec:ablation_study}
We conduct a comprehensive ablation study to investigate the effectiveness of the \textit{$\text{TOD}^3$Cap} network design. Unless specified, we utilize the 2D images as input.

\subsubsection{Effectiveness of Relation Q-Former.}
The relation modeling module is crucial for 3D dense captioning to model the intricate interactions between objects \cite{yu2023comprehensive}. Prior arts focus on modeling the relation between different specific objects with ``Graph'' \cite{chen2021scan2cap, jiao2022more, chen2021d3net} or transformer decoder \cite{wang2022spatiality, cai20223djcg, zhong2022cm3d, chen2023vote2cap}. In this section, we conduct experiments to compare different relation modules. As shown in Tab.~\ref{tab:comp_relation_modules}, the Relation Q-Former outperforms other relation modules, which is attributed to the good context awareness of the Relation Q-Former and the fact that \emph{relational graph} and \emph{transformer decoder} fail to incorporate information from BEV queries.

\begin{table*}[!t]
    \caption{Comparison of different relation modeling modules. The Relation Q-Former outperforms other relation modeling modules for its good context awareness.}
    \label{tab:comp_relation_modules}
    \setlength{\tabcolsep}{2mm}
    \centering
    % \resizebox{\linewidth}{!}{
        %
        \begin{tabular}{l|cc|cc}
            \toprule
            Relation Module   & C@0.25 & B-4@0.25 & C@0.5 & B-4@0.5\\
            \midrule
            % No                                      & -	    & -	 & -	& -    \\
            Relational Graph                        & 88.8   & 41.8	 & 82.7	& 38.4    \\
            Transformer Decoder                     & 94.9     & 44.3  & 90.0    & 41.7     \\
            Relation Q-Former (Ours)       & \textbf{96.2}     & \textbf{45.0}   & \textbf{94.1}    & \textbf{47.6}     \\
            \bottomrule
        \end{tabular}
\end{table*}

\subsubsection{Comparisons with Different Language Decoders.}
The large foundation models have been proved effective for their generalization and commonsense understanding abilities. These abilities help \textit{$\text{TOD}^3$Cap} network to well resolve long-tailed cases. To investigate the impact of the LLM decoder on \textit{$\text{TOD}^3$Cap} network, we conduct experiments on different language decoders utilized in former dense captioning methods, including S\&T \cite{vinyals2015show} and GPT2 \cite{radford2019language}, apart from LLaMA in our original setting. The results in Tab.~\ref{tab:comp_language_decoder} shows that the model with LLaMA achieves higher performance than other language decoders. This demonstrates that our network design can fully unleash the the superior language generation capabilities of large language models. Note there exist domain gaps as the original LLaMA-adapter is meant to process visual features from RGB images and language prompts, while our design processes BEV queries from multi-modal inputs and object prompts.
% To further investigate the ability of LLM decoder on captioning long-tailed objects, we specifically show results for those long-tailed objects set (named Val-Hard) in Tab.~\ref{tab:comp_language_decoder}. We can see that the LLaMA outperforms the others in a great deal. These results suggest that the performance of dense captioning on long-tailed objects can be significantly enhanced by leveraging a pretrained large language model.

\begin{table*}[!t]
    \caption{Comparison of different language decoders. The LLaMA achieves the best performance, demonstrating our network design can fully unleash the superior language generation capabilities of large language models, despite domain gaps.
    % The Val-Hard split are a human specific subset of the validation set, which only includes the long tailed objects like trash cans or brids.
    }
    \label{tab:comp_language_decoder}
    \centering
    \setlength{\tabcolsep}{2mm}
    % \resizebox{\linewidth}{!}{
        %
        \begin{tabular}{lc|cc|cc}
            \toprule
            Decoder & Adapter   & C@0.25 & B-4@0.25 & C@0.5 & B-4@0.5  \\
            \midrule
            S\&T    & Yes     & 81.2	& 32.0	& 78.6	& 29.8	 \\
            GPT2    & Yes     & 89.4	& 41.2		& 85.6	& 38.6	 \\
            LLaMA (Ours)   & Yes     & \textbf{96.2}   & \textbf{45.0}     & \textbf{94.1} & \textbf{47.6}   \\

            % \midrule
            % S\&T    & Val-Hard     & -	& -	& -	& - \\
            % GPT2    & Val-Hard     & -	& -	& -	& - \\
            % LLaMA   & Val-Hard     & - & - & - & - \\
            
            \bottomrule
        \end{tabular}
\end{table*}

\subsubsection{Impact of Different Training Strategies.}
A critical issue in our network design is the alignment between object proposal prompts with language prompts. Thus it is difficult to directly optimize the entire network from the scratch. We utilize the training strategy that divides the optimization process into several stages. We take three steps to optimize the network, (1) we pre-train the BEV-based detector on object detection task; (2) we freeze the detector weights and train the caption generation module; (3) the entire model is finetuned with a smaller learning rate. In this section, we investigate the effectiveness of the strategy we use, as shown in Tab.~\ref{tab:comp_training_strategy}.
We can see that the removal of each training phase leads to a significant performance decrease. For example, the results decrease by 8.8 on C@0.25 and by 8.8 on C@0.5 without the captioner pre-training stage. This indicates the necessities of all the pre-training.

\begin{table*}[!t]
    \caption{Comparison of different training strategies. We can see that the pretraining of detector and captioner could benefit the 3D dense captioning in outdoor scenes.}
    \label{tab:comp_training_strategy}
    \centering
    % \resizebox{\linewidth}{!}{
        %
        \setlength{\tabcolsep}{1.2mm}
        \begin{tabular}{ccc|cc|cc}
            \toprule
            Detector & Captioner & Entire Model    & C@0.25 & B-4@0.25 & C@0.5 & B-4@0.5 \\
            \midrule
                        & \checkmark   & \checkmark         & 74.2     & 39.2  & 69.5  & 37.4   \\
            \checkmark  &            & \checkmark         & 87.4    & 41.9	& 85.3 & 39.1   \\
            
            \checkmark & \checkmark & \checkmark         & \textbf{96.2}  & \textbf{45.0} & \textbf{94.1} & \textbf{47.6}   \\
            \bottomrule
        \end{tabular}

\end{table*}

\begin{table*}[!t]
    \caption{Comparison of different model scales. The ``Tuned Params'' means the parameter size that is trainable. The ``Infer-time'' represents the inference time for all 150 scenes. The smaller BEV resolution decreases the final performance while reducing the demand for memory and time cost.}
    \label{tab:comp_model_scale}
    \centering
    % \resizebox{\linewidth}{!}{
        %
        \setlength{\tabcolsep}{0.5mm}
        \begin{tabular}{c|cc|cc|cc}
            \toprule
            BEV Resolution  & Tuned Params & Infer-time & C@0.25 & B-4@0.25 & C@0.5 & B-4@0.5 \\
            \midrule
            \textit{$\text{TOD}^3$Cap}-Tiny       & 90.5M &316.1min  & 90.0 & 42.2  & 87.3  & 41.0 \\
            \textit{$\text{TOD}^3$Cap}-Small    	 & 115.4M & 331.7min  & 92.3 & 45.1  & 87.5  & 43.3 \\
            \textit{$\text{TOD}^3$Cap}       & 124.5M & 350.4min  & \textbf{96.2}  & \textbf{45.0} & \textbf{94.1} & \textbf{47.6} \\
            \bottomrule
        \end{tabular}
\end{table*}

\subsubsection{Efficiency Analysis of \textit{$\text{TOD}^3$Cap} network.}
% Involving large language models into the framework presents challenges to calculation volume and latency. 
In this study, we investigate the impact of the model scale by varying the BEV resolution, as shown in Tab.~\ref{tab:comp_model_scale}. 
% Following \cite{li2022bevformer}, we ablate the encoder scales with smaller BEV resolution, which are referred to as \textit{$\text{TOD}^3$Cap}-Small and \textit{$\text{TOD}^3$Cap}-Tiny. 
Following \cite{li2022bevformer}, our default setting is that we use Resnet101 \cite{he2016deep} as our backbone, 200*200 BEV resolution, 1600*900 input resolution and multi-scale image feautres. The \textit{$\text{TOD}^3$Cap}-small means we utilize small BEV resolution (150*150), smaller input resolution (1280*720) and single scale image features. The \textit{$\text{TOD}^3$Cap}-tiny means we utilize smaller backbone (Resnet 50), smaller BEV resolution (50*50), smaller input resolution (800*450) and single scale features.
We can see that the smaller BEV resolution decreases the final performance while reducing the demand of memory. Specifically, the \textit{$\text{TOD}^3$Cap}-Small decreases 3.9 C@0.25 and 6.6 C@0.5 and reduces 9.1M tuned parameters. The \textit{$\text{TOD}^3$Cap}-Tiny decreases 6.2 C@0.25 and 6.8 C@0.5 and reduces 34.0M tuned parameters.
% Overall, our architecture can adapt to various model scales and we hope this ablation provides insights about the accuracy-efficiency trade-off of outdoor 3D dense captioning.

\section{Conclusions}
In this study, we present the task of generating dense captions for outdoor 3D environments, utilizing both LiDAR-generated point clouds and RGB images from a panoramic camera rig. To support this task, we introduce the $\text{TOD}^3$Cap dataset, featuring 2.3 million detailed descriptions for over 64,300 outdoor objects across 850 scenes, derived from the nuScenes dataset. Our approach leverages the $\text{TOD}^3$Cap network, which employs a Relation Q-Former to understand the inter-object relationships and their contexts within a scene, and integrates with the LLaMA-Adapter for efficient caption generation without necessitating retraining of the underlying large language model. Through our contributions, we aim to facilitate advancements in outdoor 3D visual language research.
% \textbf{Limitations.}

\section*{Acknowledgement}
We would like to thank Dave Zhenyu Chen at Technical University of Munich for his valuable proofreading and insightful suggestions. We would also like to thank Lijun Zhou and the student volunteers at Li Auto for their efforts in building the \textit{$\text{TOD}^3$Cap} dataset.

\appendix
\section*{Appendix}
%------------------------------------------------------------------------------------------------%
\begin{figure}[!ht]

  \centering
  \includegraphics[width=1.\linewidth]{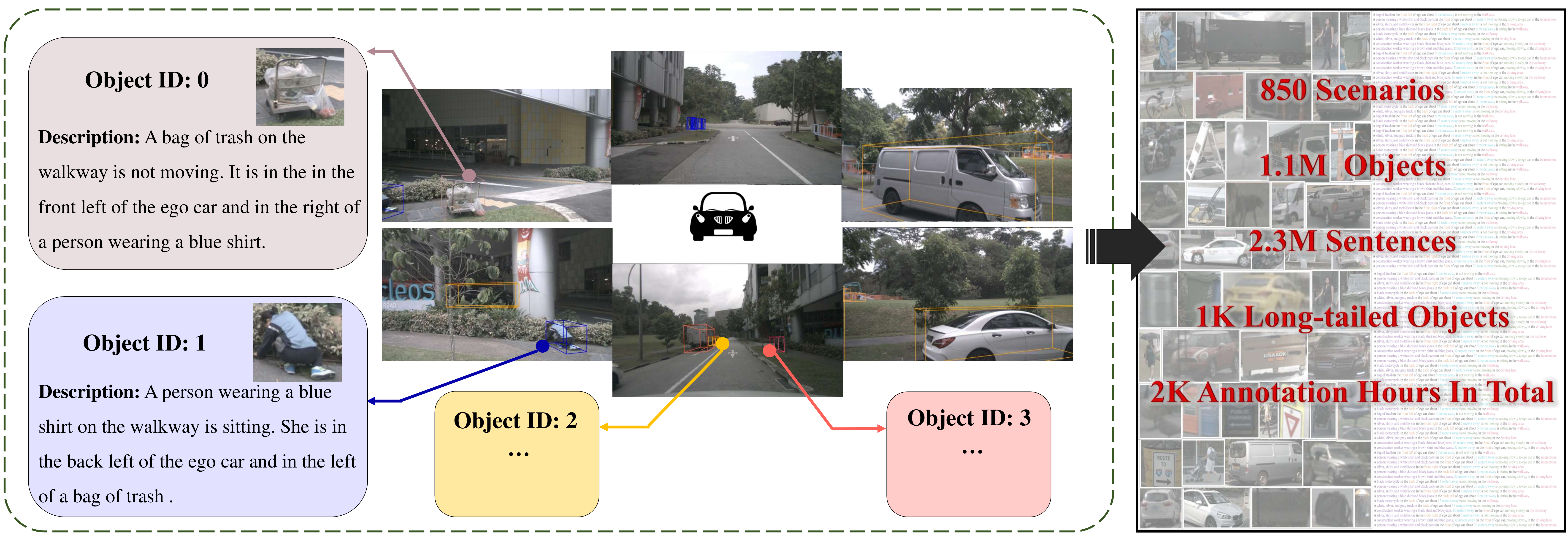}
  \vspace{-4mm}
  \caption{ \textit{$\text{TOD}^3$Cap}: Towards 3D Dense Captioning in outdoor scenes.}
  \vspace{-4mm}
  \label{fig:teaser_old}
\end{figure}
\vspace{-4mm}

\section{Annotation Details}
\begin{figure}[t]

  \centering
  \includegraphics[width=1.\linewidth]{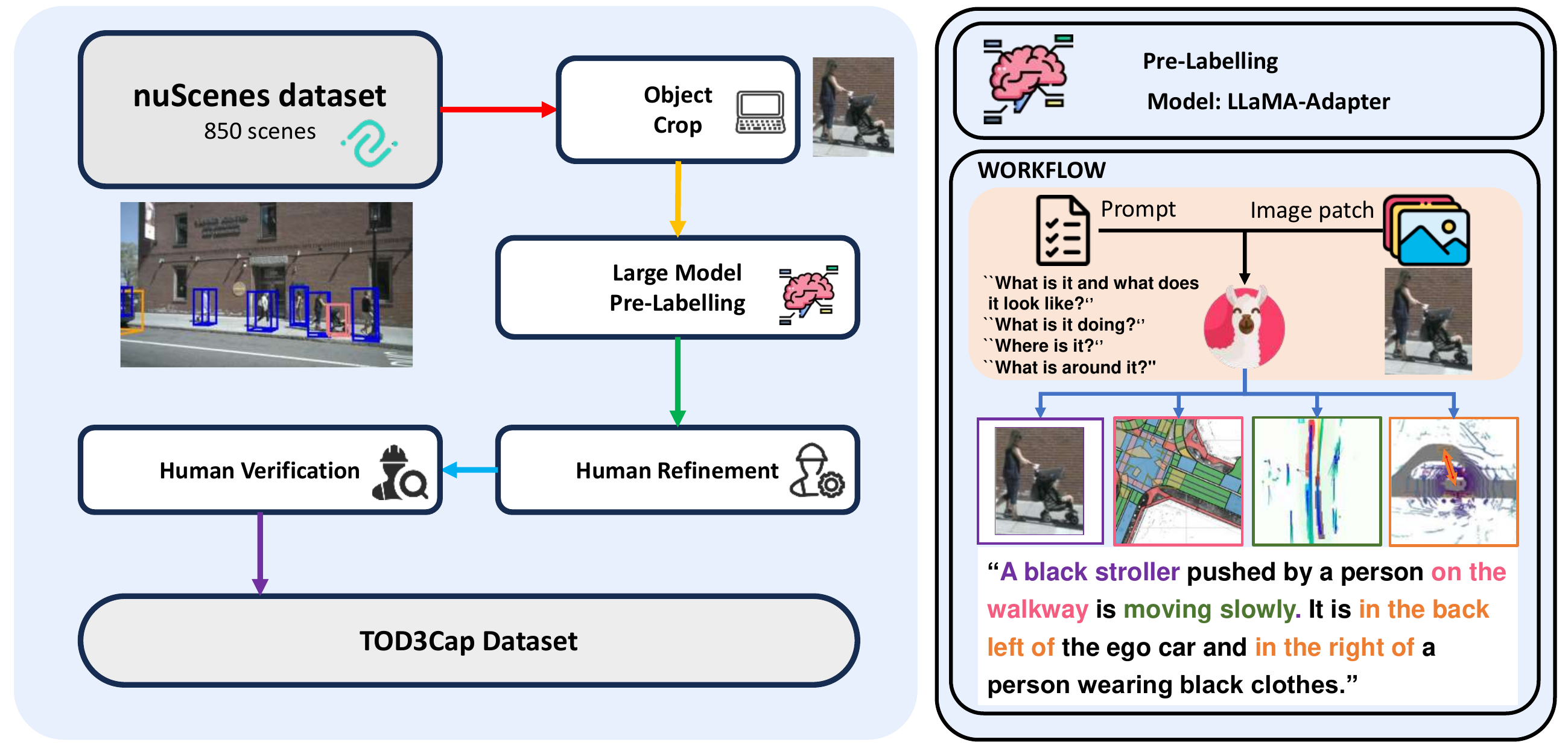}
  \vspace{2mm}
  \caption{\textbf{The Annotation Process of \textit{$\text{TOD}^3$Cap}.} Generally, there are three steps: pre-labeling, human annotation, and human verification, each of which is important to the effectiveness and efficiency of the whole process.}
  \label{fig:dataset_annotation}
\end{figure}

In this section, we provide the annotation details of \textit{$\text{TOD}^3$Cap} dataset. 
% The data annotation process involves 10 human workers and takes about 2000 hours in total.
We deploy a web-based annotation system to collect the object descriptions in outdoor scenes, which provides an interactive interface for human workers to annotate with high efficiency.
We show the overall annotation pipeline in Fig.~\ref{fig:dataset_annotation}.

\subsection{Pre-labeling}
Considering the significant progress of large foundation models in language annotation, we deploy a semi-automatic pipeline for the pre-labeling, where the human annotators would be able to choose and refine the outputs of vision-language models \cite{openai2023gpt4, zhang2023llama}.
To make the model focus on a specific object, we crop the whole camera image to an image patch that focuses on the target object and its surroundings.
The image patches are then passed as input to the vision-language models to generate descriptions. 
For the different parts of the caption (Appearance, Motion, Environment, and Relationship), we prompt the model with different questions: ``What is it and what does it look like?'', ``What is it doing?'', ``Where is it?'' and ``What is around it?''. The text outputs would be displayed when the human workers annotate the target object.

\subsection{Additional Information}
We also provide other information of the object to help the annotators analyze the properties of objects. To be mentioned, these additional information is based on the ego-car coordinate system.

\subsubsection{Viewing Direction.}
We define the viewing direction of an object $O$ as the angle between $ \overrightarrow{P_{ego} P_O} $ and the orientation of the vehicle $ R_{\text ego} $, formulated as: 
\begin{equation}
\theta=\cos ^{-1} \frac{\left(P_O - P_{ego}\right) \cdot R_{\text {ego }}}{\|\left(P_O - P_{ego} \right)\|_2\left\|R_{\text {ego }}\right\|_2},     
\end{equation}
where $P_O$ and $P_{\text ego}$ represent the position of target object and ego car, respectively. For example, when the object is in the front of the car, the $\theta$ is 0. When when the object is in the back of the car, the $\theta$ is $180\degree$.

\subsubsection{Distance.}
The distance of an object is calculated by the euclidean distance between the bounding box centers of the target and ego car, formulated as:
\begin{equation}
    d=\|P_O - P_O'\|_2, 
\end{equation}
where $P_O$ and $P_{\text ego}$ represent the position of target object and ego car.

\subsubsection{Speed.}
We also provide the speed of each object. We first differentiate the trajectory with respect to time to obtain the velocity of the object:
\begin{equation}
    V_{O} = \frac{\delta \text{trajectory}}{\delta t}, 
\end{equation}
where $\delta \text{trajectory}$ represents the position change of the ego car in $\delta t$ duration. The velocity $V_{O}$ is a vector on the global coordinate system, which is challenging for a human annotator to directly comprehend. Thus we only provide the speed of an object, defined as $||V_{O}||_2$.

% \subsubsection{Orientation.}
% When getting the velocity $V_{O}$ of an object, we can also get the orientation of it.
% The orientation is defined as the angle $\alpha$ between $\overrightarrow{O_{\text{ego}} O}$ and the velocity $V_{O}$ :

\subsection{Human Annotation}
We request the annotators to follow the following instructions:

\noindent (1) Describe the appearance of the object such as shape, color, material and so on.

\noindent (2) Describe the motion of the object, e.g., "the car is stopped" or "the man is walking".

\noindent (3) Describe the environment of the object especially its map structure, e.g., "the truck is in the carpark area" or "the child is in the crosswalk".

\noindent (4) Describe its relative position to ego car and other objects in the scene. For instance, "the car is in left of the ego car and in the back of a bus".

\begin{figure}[t]

  \centering
  \includegraphics[width=1.\linewidth]{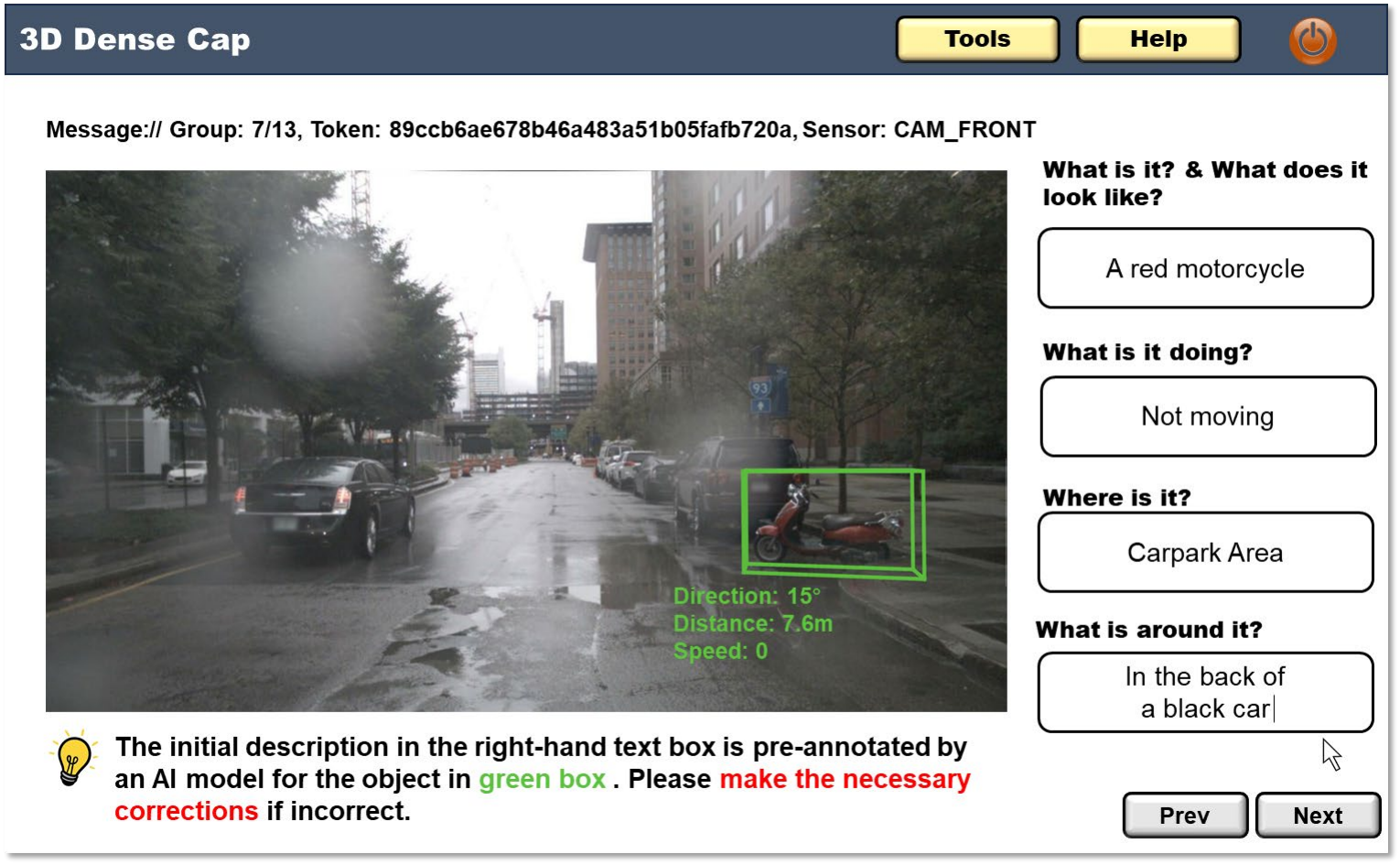}
  \caption{\textbf{The web-based UI of Annotation.}}
  \label{fig:web_ui}
\end{figure}

During annotating, annotators can refer to the information we provided previously, including the pre-labeling sentences and detailed information, as shown in Fig.~\ref{fig:web_ui}.

\subsection{Verification}
After labeling all of the four parts of the caption, we utilize AI tools \cite{openai2023gpt4} to concatenate them, referred to as ``raw captions''. Subsequently, another three human workers are employed to check the fluency and readability of the raw captions, producing ``refined captions''. The annotation will not be reserved until three annotators reach an agreement.

We provide both the raw captions and the refined captions in our dataset to ensure the descriptions are diverse and linguistically rich.

\section{Statistics}
In this section, we provide detailed statistics of the proposed \textit{$\text{TOD}^3$Cap} dataset. 

\subsection{Number of Instances and Objects}
Our \textit{$\text{TOD}^3$Cap} dataset consists of 1.1M objects and 64K instances.
Our annotations and experiments are performed on the object level. For instance, should an object be present in a scene spanning 40 frames, it will have 40 separate object descriptions to account for possible movements to different locations.

\subsection{Number of Sentences}
\begin{figure}[t]
  \centering
  \includegraphics[width=1\linewidth]{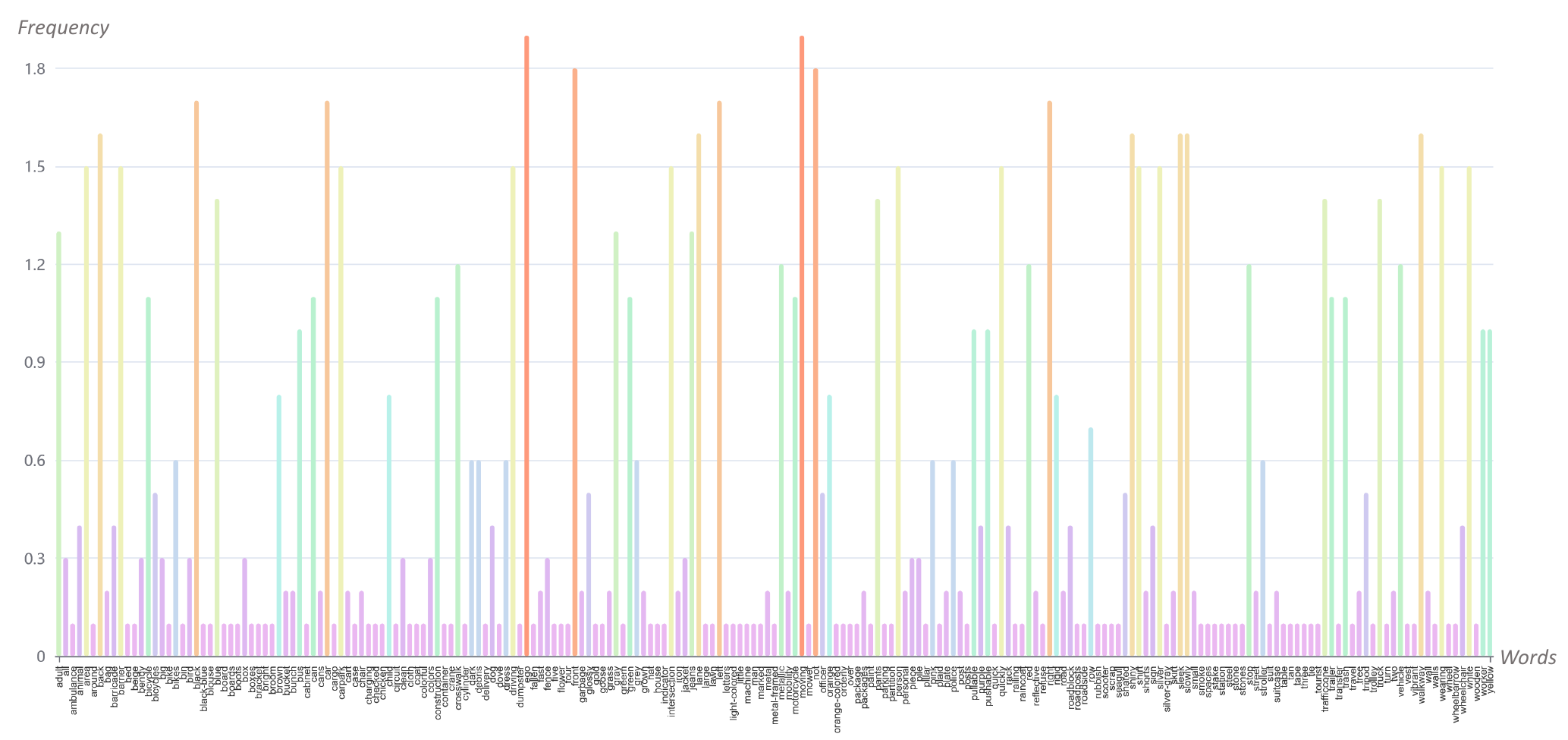}
  \vspace{-2mm}
  \caption{Statistics of word frequency.}
  %\vspace{-2mm}
  \label{fig:all_words}
  \vspace{-2mm}
\end{figure}

The total collected sentences of \textit{$\text{TOD}^3$Cap} is about 2.3M, with 1.97 sentences per object, 67.4 sentences per frame and 2705.9 sentences per scene. The total vocabulary consists of about 2k words.
In Fig.~\ref{fig:all_words}, we show the top-200 word frequency, whose frequency represents the logarithmic percentage of each word. We observe that the proportion of the first 100 words accounts for more than 90\% of the total, while some words only appear in specific scenes, like ambulance.

\begin{figure}[t]
  \centering
  \includegraphics[width=0.9\linewidth]{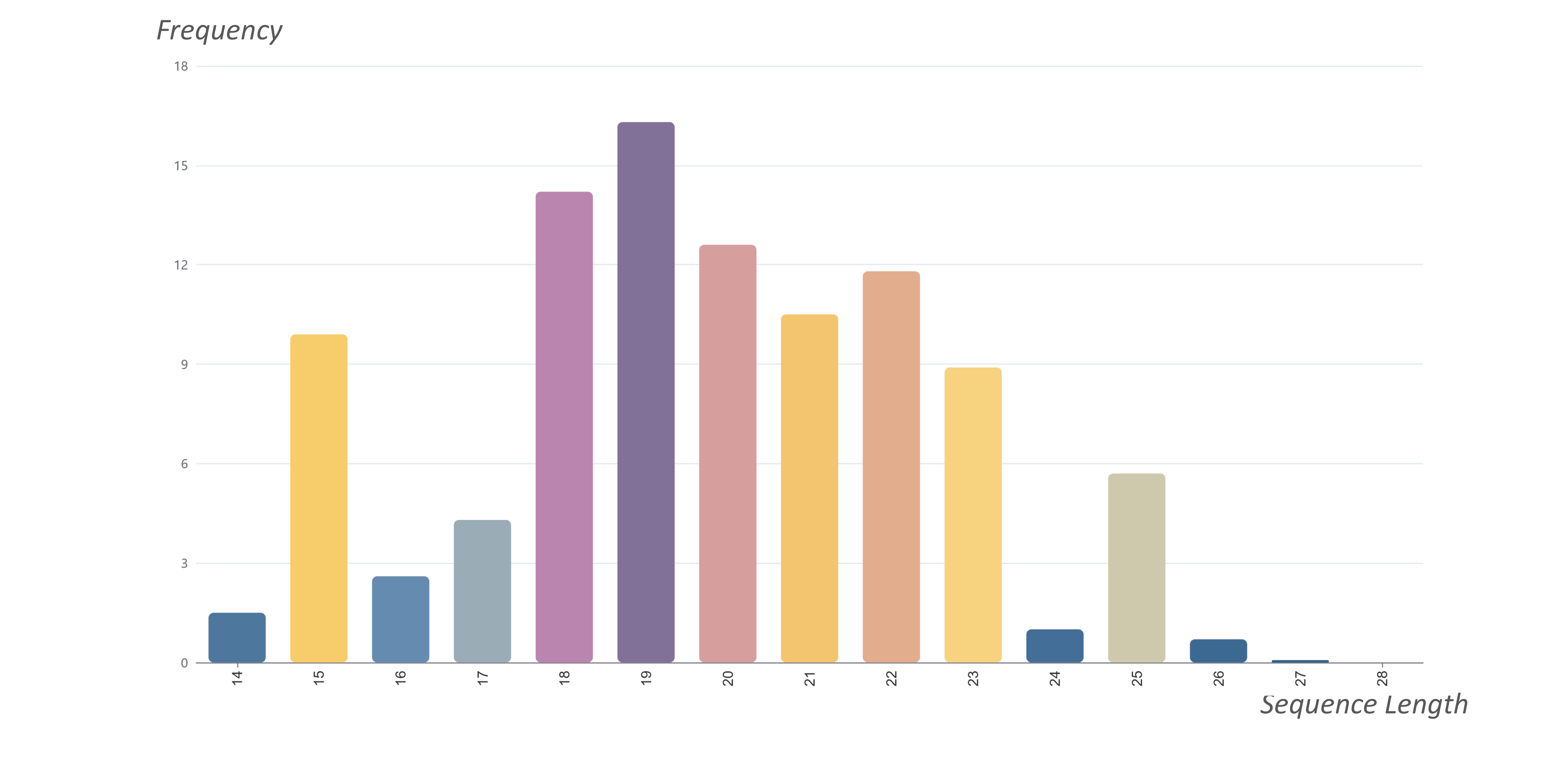}
  \vspace{-2mm}
  \caption{Statistics of sentence length frequency.}
  \label{fig:sentence_bar}
  \vspace{-2mm}
\end{figure}

\subsection{Sentence Length Frequency}
We provide the statistics of the length of all caption sentences in Fig.~\ref{fig:sentence_bar}. The sentences length ranges from 15 to 28 and the frequency represents the logarithmic percentage of each length. It can be seen that the number of words in most caption sentences is 15 to 25. 

\begin{figure}[!ht]
  \centering
  \includegraphics[width=0.9\linewidth]{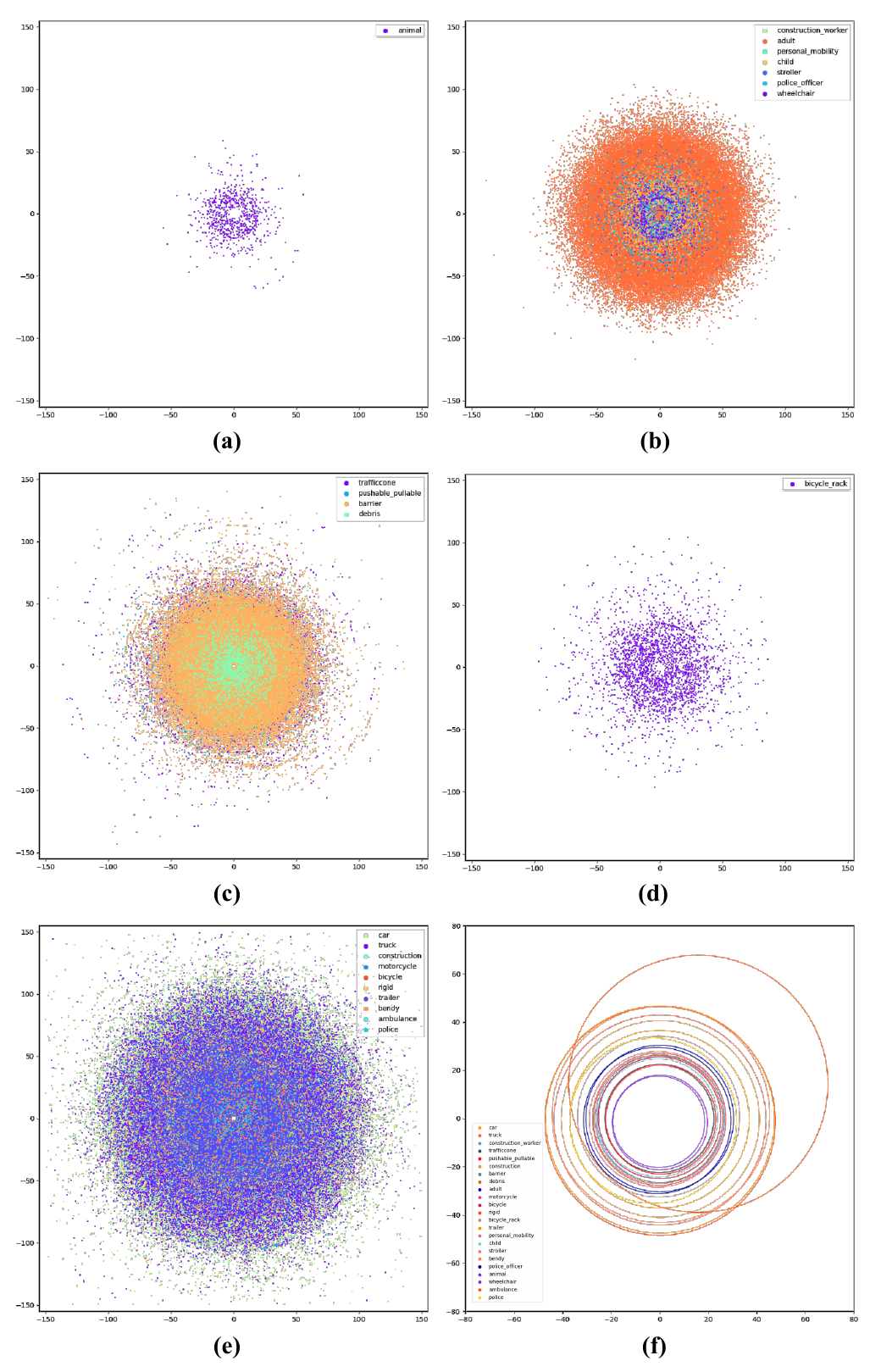}
  \vspace{-2mm}
  \caption{Visualization of objects' spatial position. (a) - (e) represents the distribution of objects' position. (f) shows the mean distribution of objects' position.}
  \label{fig:pos_cloud}
  \vspace{-2mm}
\end{figure}

\subsection{Distribution of objects' spatial position} Given the object's viewing angle and distance from the ego car, the projection of the distance in the ego car direction and perpendicular to the ego car direction can be calculated. Following this, we visualize the distribution of objects' spatial position separately according to their category, as shown in Fig.~\ref{fig:pos_cloud} where (a) - (e) represents the position distribution of animals, humans, movable objects, static objects, and vehicles relative to the ego car respectively. Besides, the visualization of the mean distribution of each kind of object is shown in Fig.~\ref{fig:pos_cloud} (f), which demonstrates the even distribution of all kinds except the ambulance. (This is because ambulances rarely appear in the dataset.)

\begin{table*}[!ht]
    \caption{Impact of Adaptation. The ``*'' represents that we replace the scene encoder with BEV encoder for adaptation. The ``-'' means we find a training collapse without adaptation. }
    \label{tab:comp_adaptation}
    \centering
    \resizebox {\linewidth}{!}{
        \begin{tabular}{c|c|cccc|cccc}
             \toprule 
             Method & Input & C@0.25 & B-4@0.25 & M@0.25 & R@0.25 & C@0.5 & B-4@0.5 & M@0.5 & R@0.5 \\

            \midrule

            \textit{$\text{TOD}^3$Cap} (Ours)    & 2D      & 96.2 & 45.0 & 34.2 & 67.4 & 94.1 & 47.6 & 33.3 & 65.4 \\

            \midrule

            Scan2Cap \cite{chen2021scan2cap}      & 3D & - & - & - & - & -  & - & - & - \\
            Scan2Cap* \cite{chen2021scan2cap}      & 3D & 50.6 &34.3 & 25.2 & 57.9 & 43.3  & 31.3 & 22.8 & 50.8 \\
            % Vote2Cap-DETR* \cite{chen2023vote2cap}          & 3D & 72.8 & 41.6 & 29.5 & 60.6 &62.6  & 35.9 & 27.4 & 55.8 \\
             % \cite{chen2023vote2cap}  & {2D+3D} & {110.1} & {48.0} & {44.4} & {67.8} & {98.4} & {46.1} & {41.3} & {65.1} \\

            \textit{$\text{TOD}^3$Cap} (Ours)  & 3D & 85.3 & 43.0 & 29.9 & 60.5 & 74.4 & 39.4 & 27.2 & 55.4       \\

            \midrule
            Scan2Cap \cite{chen2021scan2cap}      & 2D+3D & - & - & - & - & -  & - & - & - \\
            Scan2Cap* \cite{chen2021scan2cap}     & 2D+3D & 60.6 & 41.5 & 28.4 & 58.6 & 62.5 & 39.2 & 26.4 & 56.5 \\
            % X-Trans2Cap* \cite{yuan2022x}   & 2D+3D & 99.8 & 45.9 & 35.5 & 66.8 & 92.2 & 43.3 & 34.7 & 65.7 \\
            % Vote2Cap-DETR* \cite{chen2023vote2cap}  & {2D+3D} & {110.1} & {48.0} & {44.4} & {67.8} & {98.4} & {46.1} & {41.3} & {65.1} \\

            \textit{$\text{TOD}^3$Cap} network (Ours)  & 2D+3D     & \textbf{120.3} & \textbf{51.5} & \textbf{45.1} & \textbf{70.1} & \textbf{108.0} & \textbf{50.2} & \textbf{48.9} & \textbf{69.2} \\
            \bottomrule
        \end{tabular}
    }
\end{table*}

\begin{figure}[!ht]
  \centering
  \includegraphics[width=0.9\linewidth]{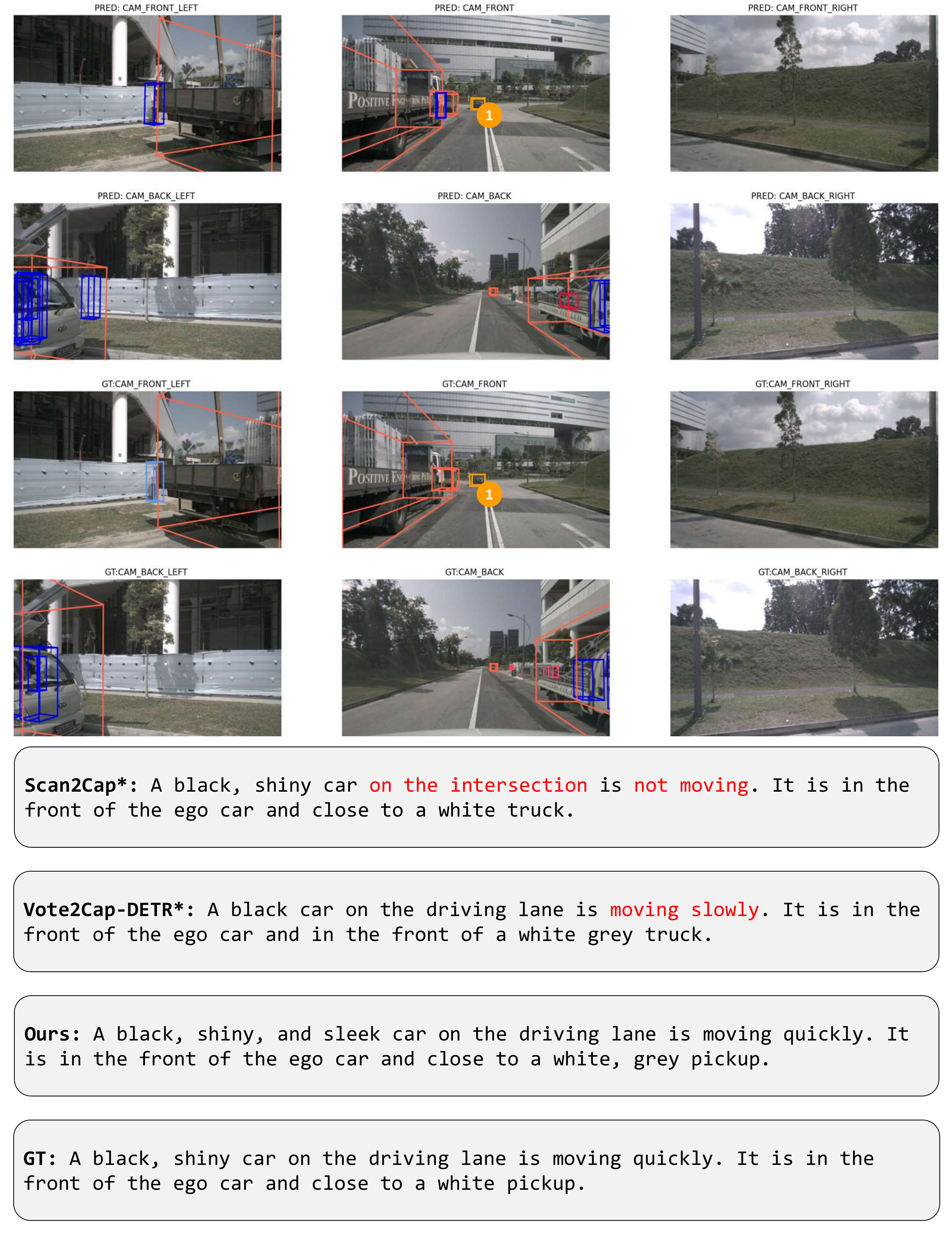}
  \vspace{-2mm}
  \caption{Visualization of different methods.}
  \label{fig:sup_vis_method}
  \vspace{-2mm}
\end{figure}

\begin{figure}[!ht]
  \centering
  \includegraphics[width=0.9\linewidth]{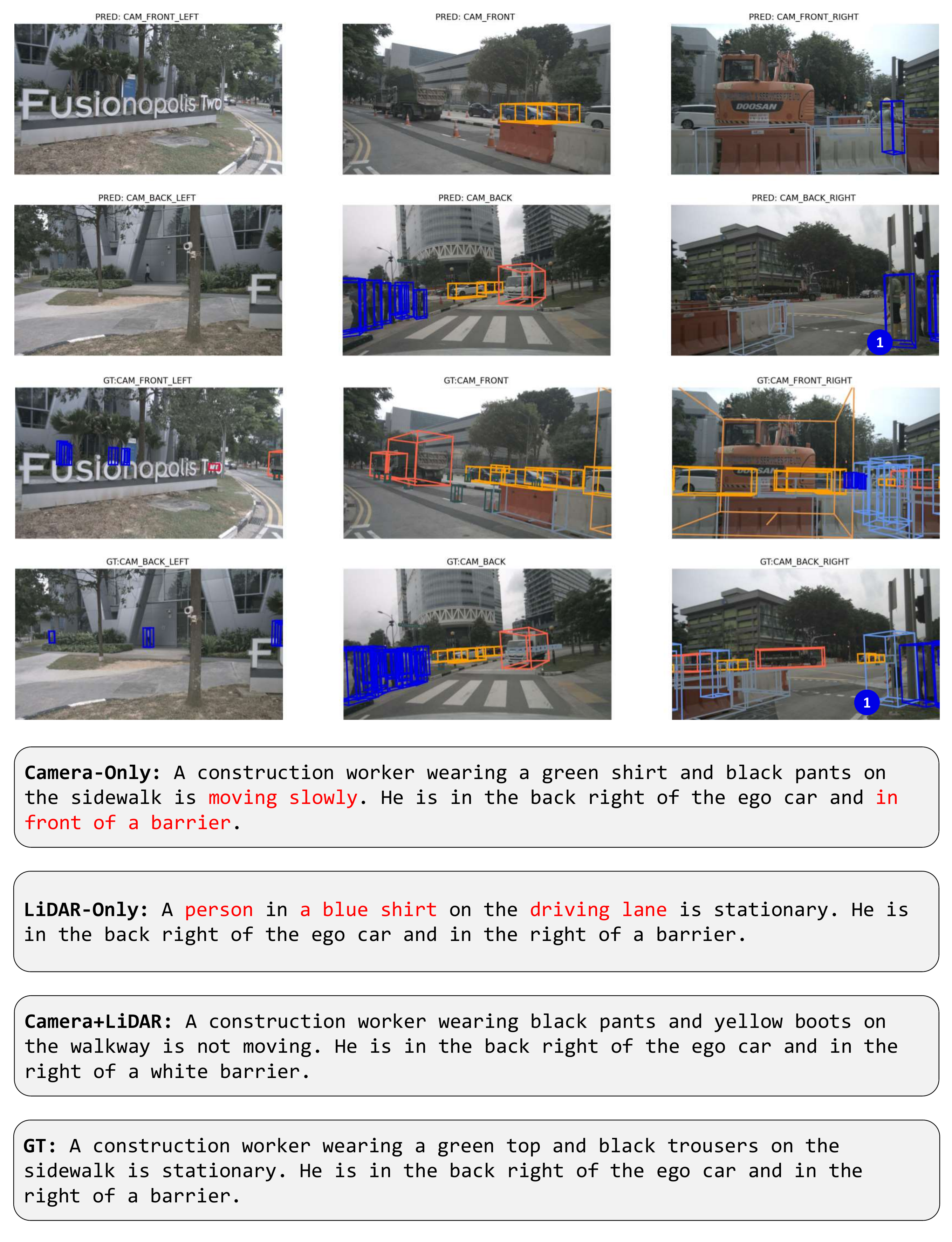}
  \vspace{-2mm}
  \caption{Visualization of \textit{$\text{TOD}^3$Cap} network with different input modalities.}
  \label{fig:sup_vis_modal}
  \vspace{-2mm}
\end{figure}

\section{More Experiments}
In this section, we introduce some additional experiments to investigate the effectiveness of the \textit{$\text{TOD}^3$Cap} network.

\subsection{Impact of Adaptation}
We find that directly applying the previous arts to our dataset without adaptation would lead to a training failure (shown in Tab.~\ref{tab:comp_adaptation}), illustrating the importance of adaptation.

\section{More Qualitative Results}
In this section, we show more visualization results of different methods.

\subsection{Comparisons among different methods}
We showcase the qualitative results of different methods in Fig.~\ref{fig:sup_vis_method}. We observe that all of the methods can perceive the appearance of objects well. However, our \textit{$\text{TOD}^3$Cap} can also get good results on the captions of objects' environment and motion.

\subsection{Qualitative Results with Different Input Modalities}
We showcase qualitative results with different input modalities in Fig.~\ref{fig:sup_vis_modal}. We can see that the model with camera-only input can successfully capture the appearance (construction worker wearing a green shirt and black pants) of the object and its environment (sidewalk). However, it misunderstands the motion state of the target and its relationship with surrounding objects. This may be because it is difficult to capture distance information of objects solely based on images, leading to its misunderstanding of the object's spatial position and thus getting the wrong motion and wrong spatial relation.

In contrast, the model with LiDAR-only input can obtain relatively accurate location information. Although it can infer the category of an object from its shape (person), it cannot obtain information about its appearance and its environment, resulting in a wrong caption.

\section{Applications}
The dataset we propose has diverse downstream applications, such as 2D/3D visual question answering, image synthesis, or autonomous driving planning.

\subsection{2D/3D Question Answering}
\label{subsection:2D Question Answering}
\begin{figure}[t]
  \centering
  \includegraphics[width=0.9\linewidth]{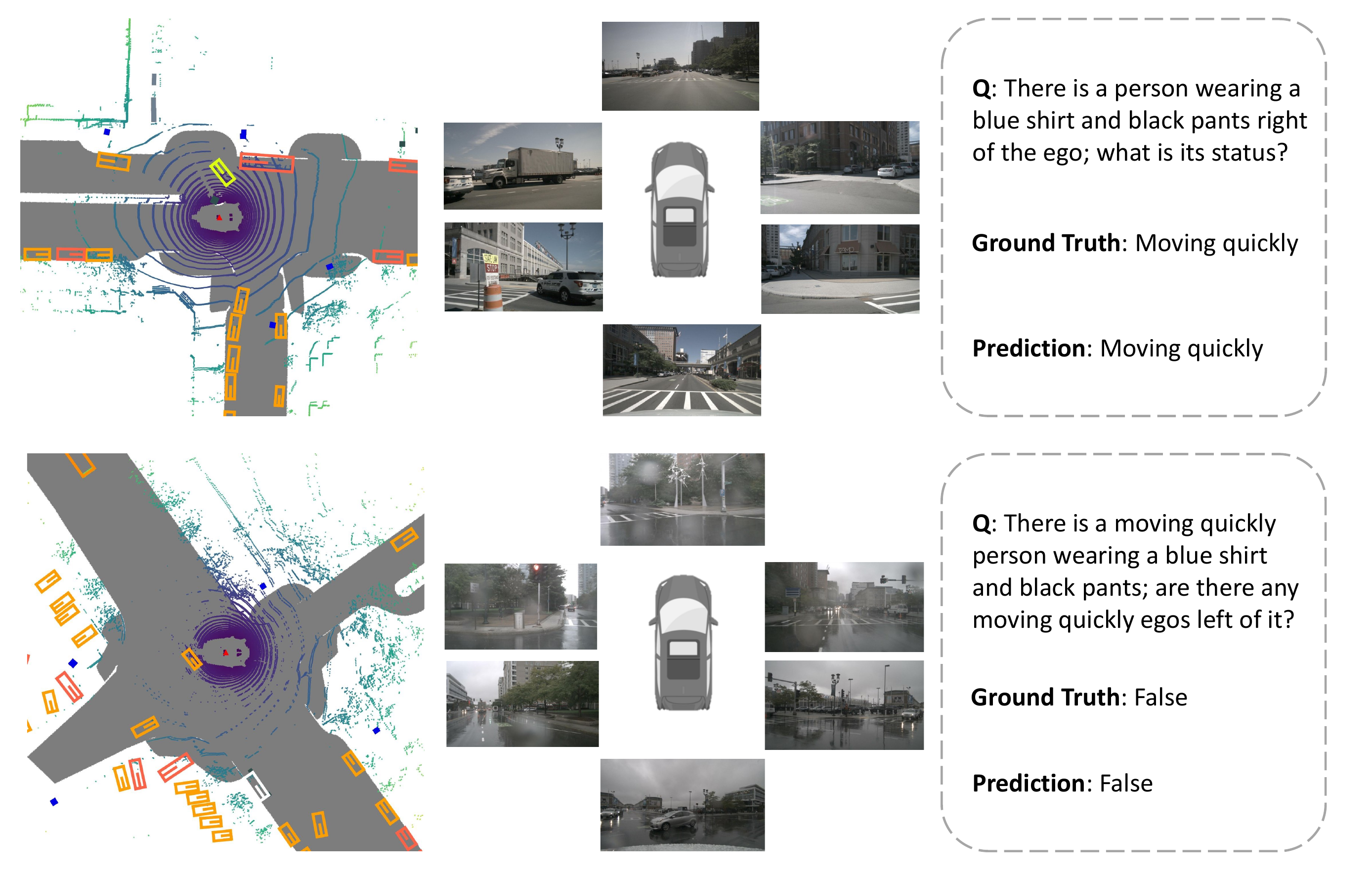}
  \vspace{-2mm}
  \caption{Qualitative results of driving QA.}
  %\vspace{-2mm}
  \label{fig:testqa}
\end{figure}

Based on \textit{$\text{TOD}^3$Cap}, we further develop a visual question answering (VQA) benchmark in autonomous driving scenarios regarded as driving QA. According to previous captions, we design various kinds of the question with structured data generated from dense captioning of nuscenes, encompassing existence, counting, query-object, query-status, and comparison, inspired by CLEVR benchmark \cite{johnson2017clevr}. For instance, the template for questions of the existence type can be expressed as follows: "Are there any $<A2>$ $<O2>$ to the $<R>$ of the $<A1>$ $<O1>$?," where $<A2>$, $<O2>$, and $<R>$ correspond to distinct parameter types, specifically appearance, object category, and relationships. Eventually, we obtain a total of 468K question-answer pairs on 850 scenes from the annotated nuScenes training and validation split, with 385K pairs for training and 83K for testing.
Furthermore, we establish a simple model following \cite{zhang2023llama} and evaluate the performance of the visual question answering on driving QA.
% To assess the performance of the QA task, we employ Top-1 accuracy as our evaluation metric, in line with the common practice in various VQA studies \cite{2015VQA}. Additionally, we incorporate sentence evaluation metrics commonly utilized in image captioning models since certain questions have multiple valid answer expressions. Specifically, we integrate BLEU \cite{papineni2002bleu}, ROUGE \cite{lin2004rouge}, and METEOR \cite{banerjee2005meteor} metrics to evaluate the answer matching.
We show some results in Tab.~\ref{table:comp_vqa_2d} and Fig.~\ref{fig:testqa}.

\vspace{-2mm}
\begin{table*}[!ht]
\begin{center}
	\footnotesize\begin{tabular}{lccccccc}
        \toprule
Model     & Accuracy & BLEU-1 & BLEU-2& BLEU-3 & BLEU-4 & ROUGE & METEOR  \\
\midrule

Ours (single-view)   & 46.02 & 56.67 & 19.38 & \textbf{0.20} &0.02 &51.14 & 56.68 \\
Ours (multi-view) & \textbf{47.77} & \textbf{58.45} & \textbf{19.50} & 0.19 &\textbf{0.02} &\textbf{52.97} & \textbf{58.46} \\

\bottomrule
	\end{tabular}
 \vspace{2mm}
    \caption{
Performance comparison of 2D question answering results. The ``multi-view'' denotes that we utilize multi-view images as input and the ``single-view'' means we only utilize single-view images as input.
    }
    \label{table:comp_vqa_2d}
\end{center}
%\vspace{-4mm}
\end{table*}

\begin{figure}[t]
  \centering
  \includegraphics[width=1.0\linewidth]{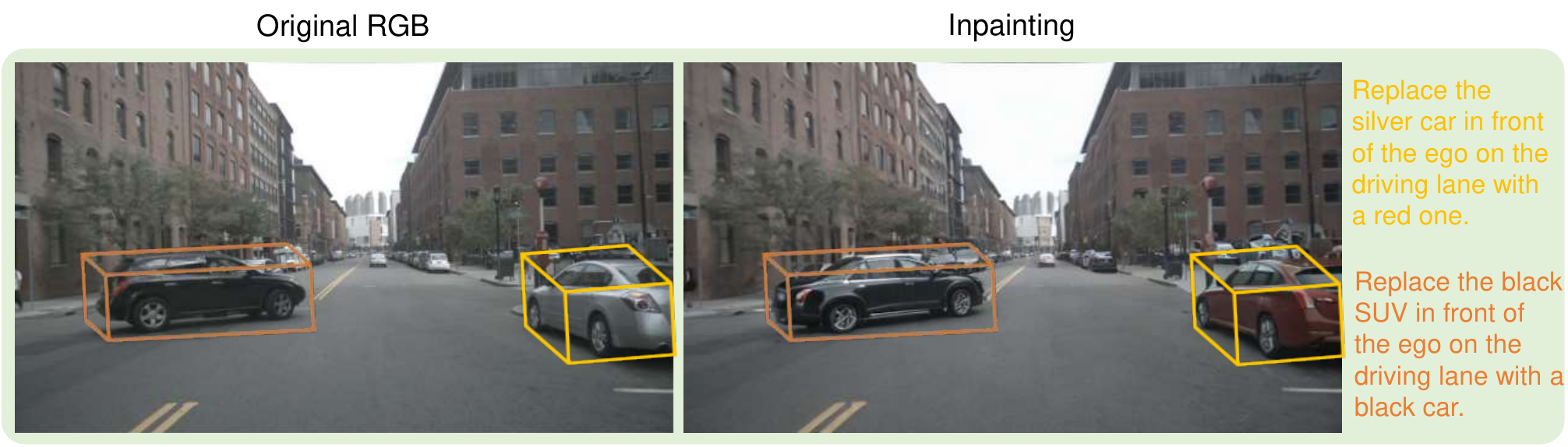}
  % \vspace{-2mm}
  \caption{Qualitative results of image synthesis.}
  %\vspace{-2mm}
  \label{fig:inpainting}
\end{figure}

\subsection{Image Synthesis}
Our detailed object-level descriptions also benefit the image synthesis like the inpainting. Following \cite{yang2023bevcontrol}, we replace the original label condition with the text condition, with results shown in Fig.~\ref{fig:inpainting}. We can see that our caption can provide detailed guidance for the generated images.

\subsection{LLMDriverAgent}
\label{subsection:LLMDriverAgent}
\begin{figure}
  \centering
  \includegraphics[width=1.0\linewidth]{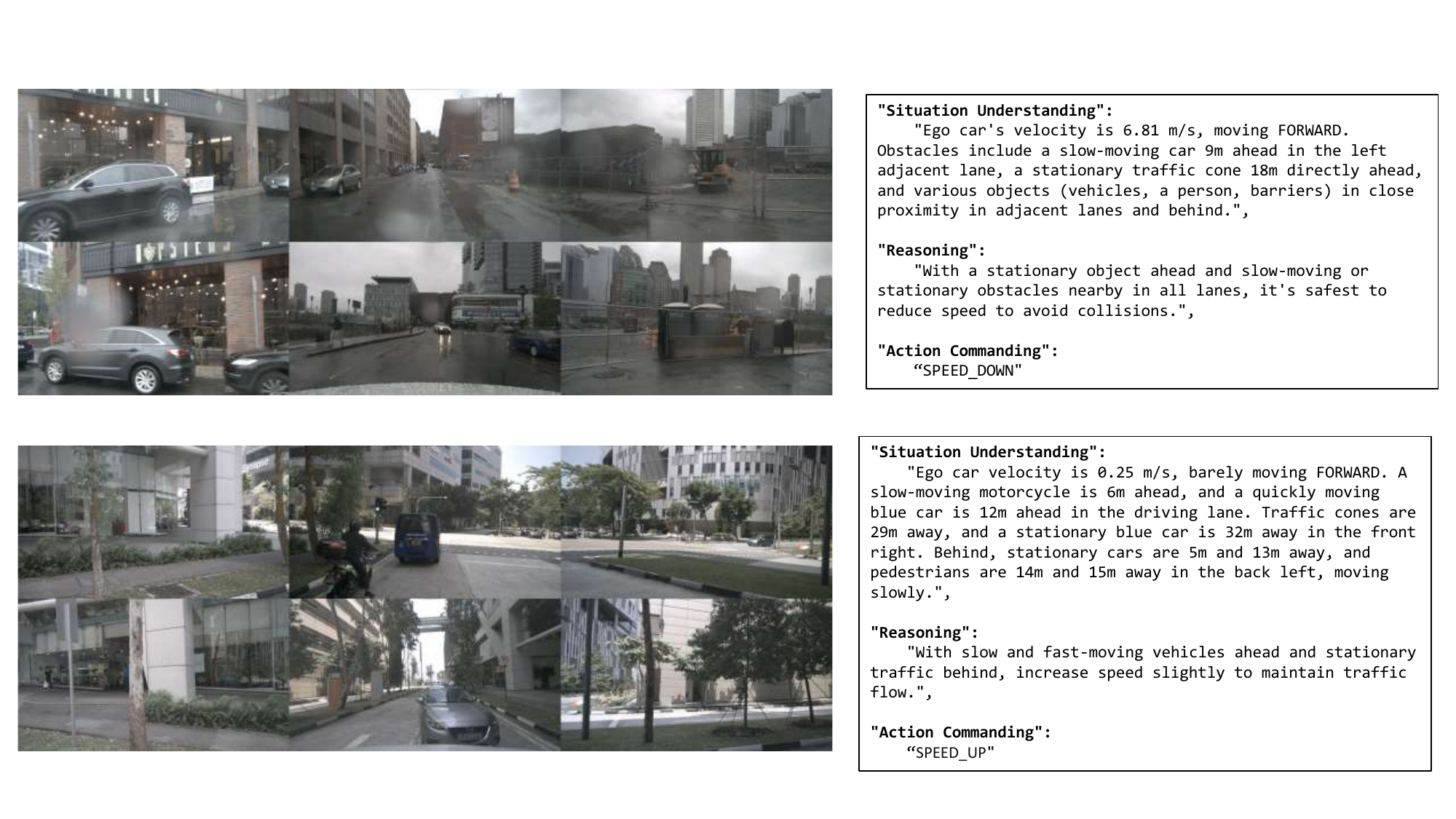}
  % \vspace{-4mm}
  \caption{More example outputs of LLM driver agent.}
  \label{fig:driving_agent_output}
  % \vspace{-5mm}
\end{figure}

\begin{figure}
  \centering
  \includegraphics[width=1\linewidth]{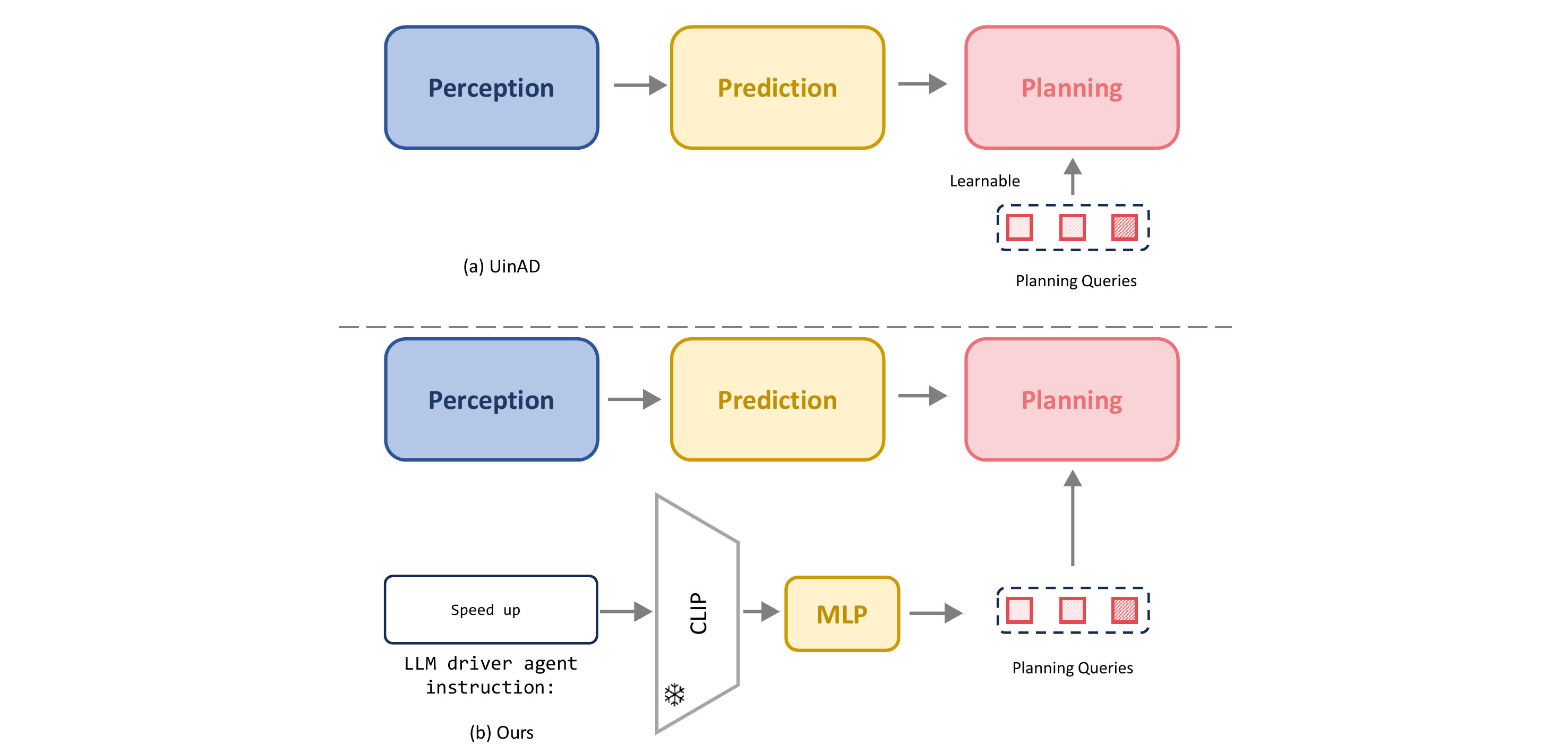}
  % \vspace{-2mm}
  \caption{The architecture of UniAD + LLMDriverAgent.}
  \label{fig:llm_drive}
\end{figure}

\begin{figure}
  \centering
  \includegraphics[width=1\linewidth]{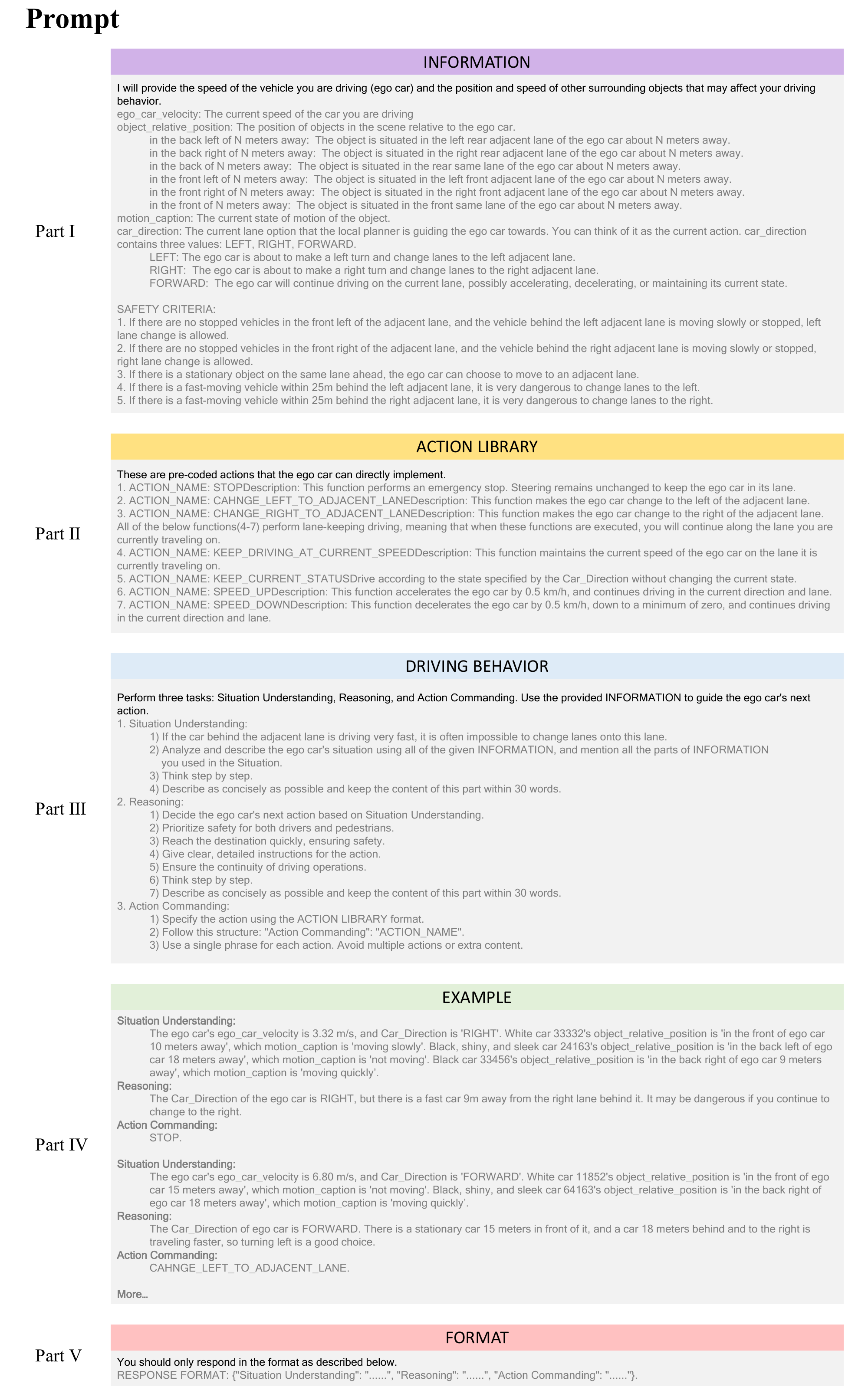}
  \vspace{-2mm}
  \caption{Prompt input to LLM driver agent.}
  \label{fig:planning-prompt}
\end{figure}

\begin{figure}
  \centering
  \includegraphics[width=1\linewidth]{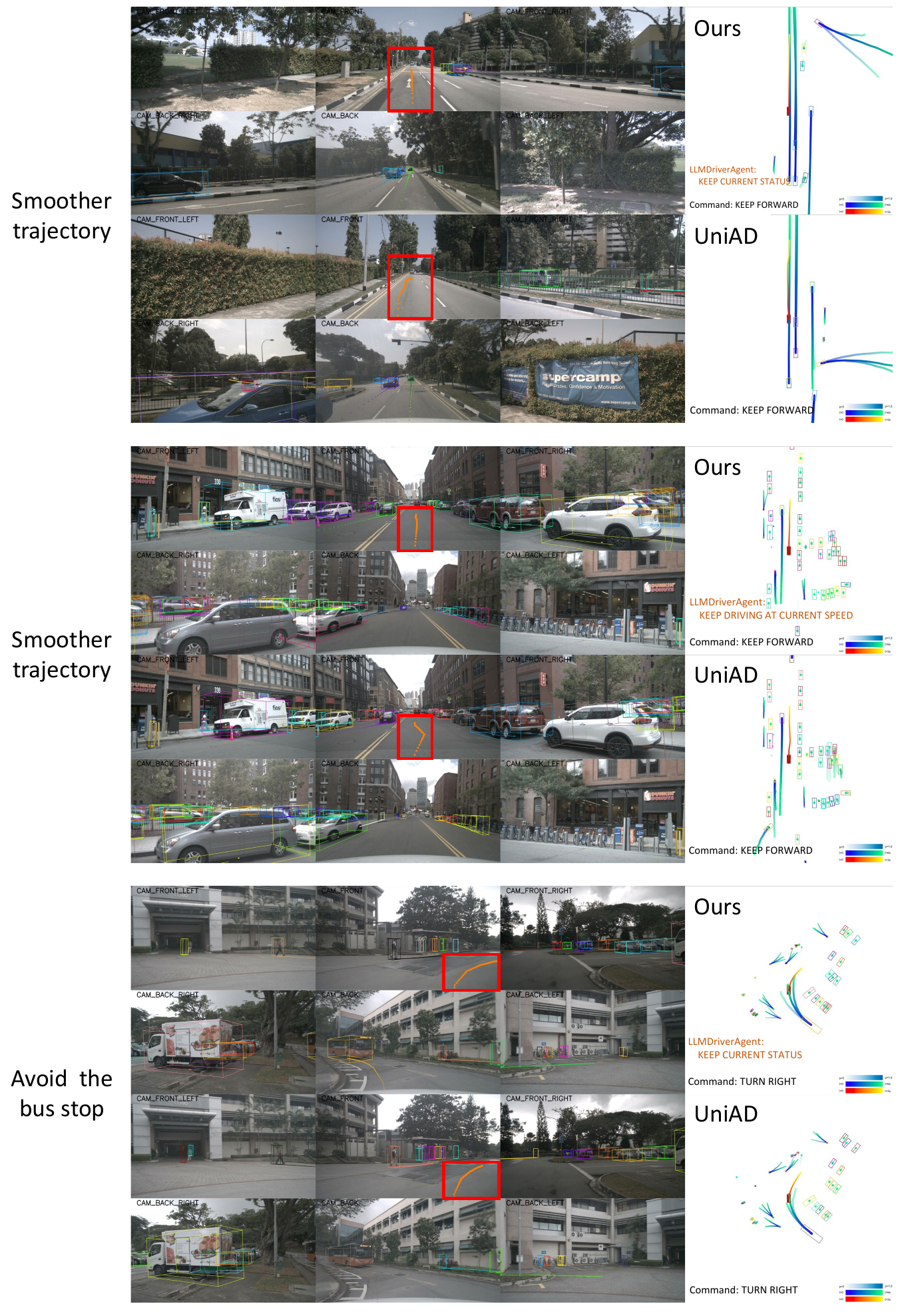}
  \vspace{-2mm}
  \caption{Qualitative results of UniAD and our proposed method. Red boxes show the comparison between UniAD's trajectories and ours. In cases 1 and 2, our predicted trajectories are smoother as our method is more aware of the interaction with the environment and vehicles. In case 3, our predicted trajectory successfully avoids the bus stop where UniAD fails.}
  \label{fig:compare_uniad}
\end{figure}

\begin{figure}
  \centering
  \includegraphics[width=0.8\linewidth]{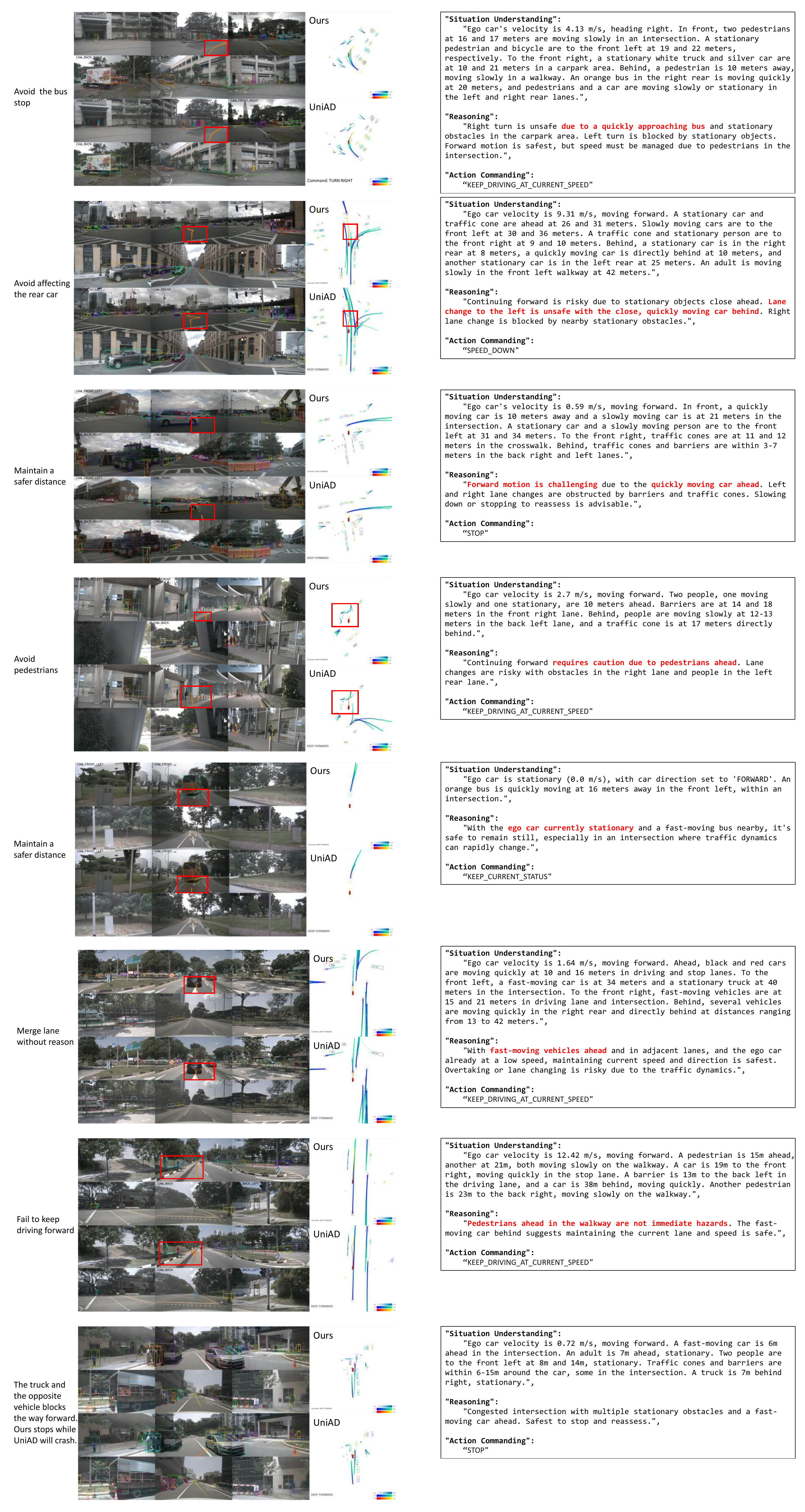}
  \vspace{-2mm}
  \caption{{More qualitative results of UniAD and our proposed method. Red boxes show the comparison between UniAD's trajectories and ours. The right column shows the output of the LLM agent.}}
  \label{fig:llm_drive_new}
\end{figure}

In order to explore the effectiveness of large language model (LLM) on autonomous vehicle planning and decision, following \cite{fu2023drive}, we generate driving prompts as shown in Fig. \ref{fig:planning-prompt}. These prompts are then fed into ChatGPT to obtain autonomous driving instructions, creating what we call an LLMDriverAgent. We use the output of LLMDriverAgent or say driving instruction, shown in Fig.~\ref{fig:driving_agent_output}, as additional inputs to UniAD (the framework is shown in Fig.~\ref{fig:llm_drive}). The results, shown in Fig.~\ref{fig:compare_uniad}, \ref{fig:llm_drive_new} and Tab.~\ref{tab:comp_uniad} demonstrate the effectiveness of natural language commands on autonomous vehicle planning.

\begin{table*}[!ht]
    \centering
    \resizebox{\linewidth}{!}{
        \begin{tabular}{l|c|cccc|cccc}
            \toprule
            Method  & Input & L2 1s & L2 2s & L2 3s & Avg. & Collision 1s   & Collision 2s  & Collision 3s   & Avg. \\

            \midrule

            UniAD \cite{hu2023planning}   & 2D    & \textbf{0.48}  & \textbf{0.96}  & 1.65  & \textbf{1.03} & 0.05           & 0.17          & 0.71          & 0.31 \\

            Ours    &2D     & {0.53} & {0.98} & \textbf{1.64}  & {1.05}  & \textbf{0.03} & \textbf{0.12} & \textbf{0.55} & \textbf{0.23} \\

            \bottomrule
        \end{tabular}
    }
    \vspace{5mm}
    \caption{Comparison of planning results on nuScenes. With language suggestions, the results of UniAD have been greatly improved, especially on collision rate despite the l2 loss has been slightly improved.}
    \label{tab:comp_uniad}
\end{table*}
\vspace{10mm}

% \vspace{-4mm}

\bibliographystyle{splncs04}
\bibliography{main}
\end{document}